%% file: main.tex
\documentclass{article}

\usepackage{iclr2027_conference,times}

\usepackage{amsmath, amssymb}
\usepackage{graphicx}
\usepackage{booktabs}
\usepackage{hyperref}
\usepackage{microtype}
\usepackage{xcolor}

\graphicspath{{figs/}}

\title{The Circuit Basis of Metacognitive Failure in Language Models: Interference and Hallucination}

\author{%
  Qiyao Liang\\
  Massachusetts Institute of Technology\\
  \texttt{qiyao@mit.edu}
  \And
  Risto Miikkulainen\\
  University of Texas at Austin, Cognizant\\
  \texttt{risto@cs.utexas.edu}
  \And
  Ila Fiete\\
  Massachusetts Institute of Technology\\
  \texttt{fiete@mit.edu}
}

\iclrfinalcopy

\begin{document}

\maketitle

\begin{abstract}
  Language models often answer with high confidence, as measured by the
  entropy of the output distribution, even when they are wrong. We study two
  situations that lead to such errors, \emph{knowledge conflict}, where the
  context contradicts a fact stored in the weights and the two sources
  \emph{interfere}, and \emph{missing
  knowledge}, where the queried fact was never learned and the model
  \emph{hallucinates}. To control exactly what the model has learned, we
  fine-tune it on synthetic facts that map arbitrary entity identifiers to
  random codes, installed through low-rank adapters (LoRA) on the multi-layer
  perceptron (MLP), query--key (QK), or value--output (VO) components of the base transformer. Our
  analysis suggests that MLP is the main substrate for storing facts in
  \emph{basins}, regions of representation space into which a fact's
  representations converge, while QK steers generation toward a basin or the
  context, and VO writes the retrieved content into the residual stream.
  \emph{Under knowledge conflict}, the form of interference depends on how
  the fact is learned. A fact
  learned from a single phrasing yields to the context, with each adapter
  leaving an error pattern consistent with its circuit role. A fact learned from many phrasings overrides the
  context, and under contradiction the answer starts on the stored code and
  drifts away from it.
  \emph{Under missing knowledge}, the hidden state reaches no basin, yet we found that all
  adapter placements produce confident hallucinations, with the VO adapter producing the most out of all.
  In both situations, the output entropy reflects how sharply the model
  commits rather than whether the answer is correct, even though the model's
  representations still separate correct answers from errors. Across 12 pretrained
  models, the overall hallucination rate falls with scale, but a growing
  fraction of the remaining hallucinations are confident.
  In the models we study, the gap between confidence and correctness does not
  close with scale, and flagging these errors requires confidence that reflects where the
  answer comes from, rather than how sharply the model commits to it.
\end{abstract}

\input{sections/1_intro}
\input{sections/2_task}
\input{sections/3_circuit}
\input{sections/4_failures}
\input{sections/5_confidence}
\input{sections/6_related}
\input{sections/7_discussion}

\bibliographystyle{iclr2027_conference}
\bibliography{refs}

\newpage
\section*{Appendix}
\appendix
\input{appendix}

\end{document}

%% file: sections/1_intro.tex
\section{Introduction}
\label{sec:intro}

Language models often answer with high confidence even when they are wrong.
We study two situations that lead to such errors. In \emph{knowledge
conflict}, the context contradicts a fact stored in the model's weights
\citep{longpre2021entity,neeman2023disentqa}. The two sources can then
disrupt each other. The model may ignore the context, misread it, or answer
from neither source. We call this \emph{interference}, after the
corresponding phenomenon in human memory \citep{underwood1957interference}.
In \emph{missing knowledge}, the model is asked about something it never
learned and answers fluently anyway. This is a \emph{hallucination}
\citep{ji2023survey,zhang2023siren}. In both situations, the answer can look
just as sensible as a correctly recalled fact. Such errors are usually
attributed to miscalibration, which post-hoc methods such as temperature
scaling could in principle fix \citep{guo2017calibration}. This explanation
assumes that the model's confidence carries the right information on the
wrong scale. However, it has not been systematically studied where the
information that separates a correct answer from an error lives in the
model, or why the model's confidence does not reflect it.

\begin{figure}[t]
  \centering
  \includegraphics[width=\textwidth]{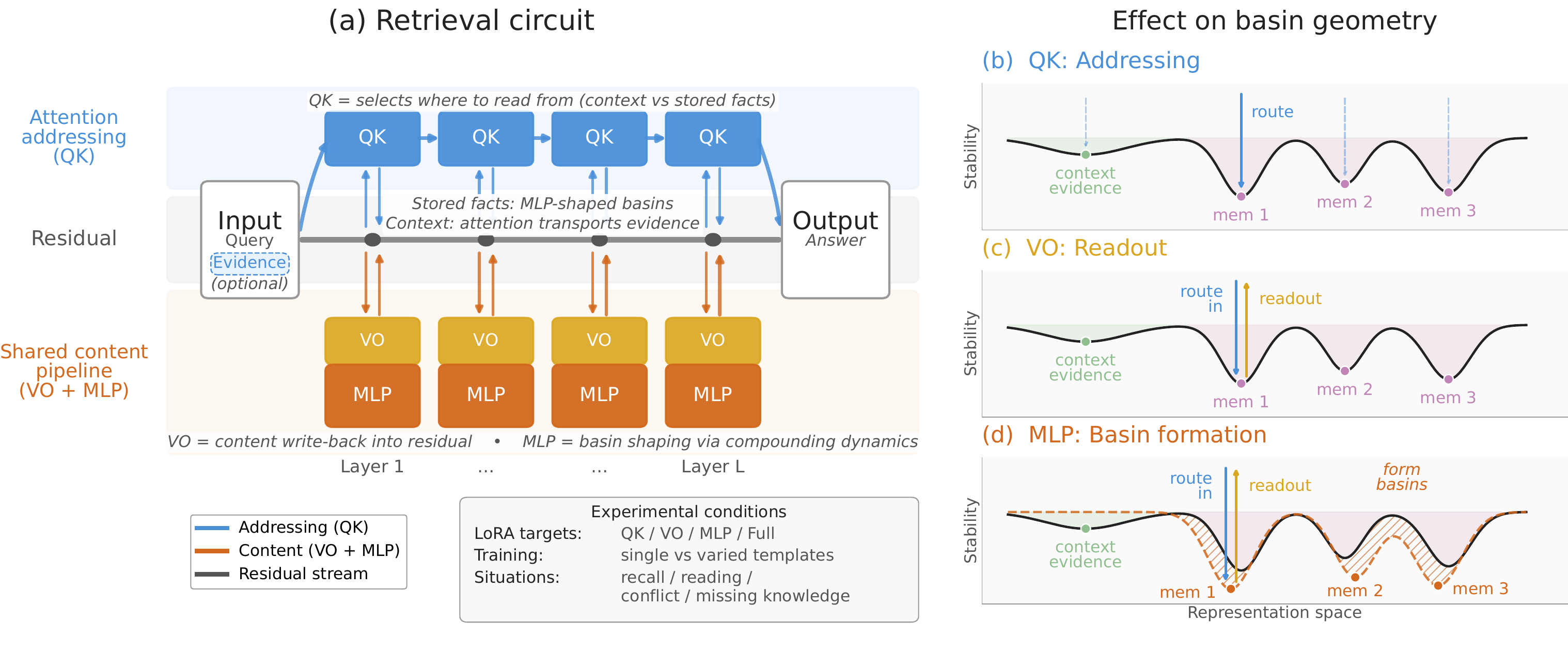}
  \caption{\textbf{The retrieval circuit and its basin geometry.} \emph{(a)}~The
    circuit behind contribution 1: an addressing mechanism (QK, blue) that
    steers the trajectory toward the context or toward stored content, and a
    content pipeline (VO and MLP, orange) that writes content into the residual
    stream and shapes the state. \emph{(b--d)}~The effect of placing an
    adapter on each component, behind contributions 2 and 3. QK changes which
    basin is selected without reshaping any, VO changes what is read out
    within the selected basin, and MLP reshapes the basins themselves. The
    output entropy depends only on how sharply the final state commits, and does
    not read whether a basin was reached or which source won.}
  \label{fig:schematic}
\end{figure}

Studying this requires knowing exactly what a model does and does not know.
Natural language does not allow this. A fact may have been seen during
pretraining, and a correct answer could come from recall, inference, or a
lucky guess. We therefore teach a pretrained LLM (Qwen2.5-3B-Instruct)
1{,}600 associations between arbitrary entity identifiers and random
five-digit codes, which it cannot infer or recover from pretraining. Our
design lets us vary three things independently: i) \emph{what} the model
knows, which we control exactly; ii) \emph{how robustly} it knows it, by
training on a single phrasing or on many; and iii) \emph{where} the knowledge
is installed, by placing low-rank adapters (LoRA) \citep{hu2021lora} on QK,
VO, or MLP. We describe the results in terms of the generation trajectory,
the path the hidden state follows through representation space as the model
generates. When a fact is learned from many phrasings, these paths converge
into a common region, which we call a \emph{basin}
(Section~\ref{sec:region}).

\paragraph{Contributions.} Our setting lets us make four contributions to the
understanding of memory and metacognition in LLMs. The first identifies the
roles of the circuit components. The next two show how these roles produce
errors in each situation, and the last shows how the problem changes with
scale:
\begin{enumerate}
  \item \textbf{QK, VO, and MLP play distinct roles in retrieval
  (Section~\ref{sec:circuit}).} MLP is the main substrate for storing facts
  in basins, consistent with its known role as a key--value memory
  \citep{geva-etal-2021-transformer,meng2022locating}. The evidence suggests
  that QK steers the trajectory toward a basin or toward the context, and
  that VO writes the retrieved content into the residual stream. We infer
  these roles from the Jacobians of the pretrained model and support them
  with adapter placement.
  \item \textbf{Under knowledge conflict, the form of interference
  depends on how the fact is stored (Section~\ref{sec:conflict}).} A fact
  learned from a single phrasing has a shallow basin, so the model answers
  from the context. When the adapter reshapes the basins (MLP or Full),
  however, the model copies the first digits from the context correctly but
  errs on the last ones, as if the stored codes take over partway through the
  answer. A fact learned from many phrasings has a basin so deep that the
  model stops reading the context, which we call \emph{context deafness}.
  Under a direct contradiction, the answer then starts on the stored code and
  drifts into one that matches neither source. In all these cases, entropy
  stays low.
  \item \textbf{Under missing knowledge, every adapter placement produces
  confident hallucinations (Section~\ref{sec:hallucination}).} Without an
  adapter, the base model declines to answer about unseen entities. With any
  adapter, it produces a confident code, even though its hidden state lies
  outside every basin. VO makes these hallucinations the most confident,
  while MLP and Full bias them toward memorized codes.
  \item \textbf{The fraction of hallucinations that are confident grows with scale
  (Section~\ref{sec:scale}).} Across 12 pretrained LLMs from four families,
  larger models hallucinate less often, but a larger share of their remaining
  hallucinations are confident. A relation derived from the output logits
  predicts this trend.
\end{enumerate}

These observations reveal the model's lack of metacognition, that is, its
inability to know what it knows. The model's confidence reflects how sharply
it commits to an answer, not whether the answer is correct. Yet its hidden
states still contain signals that distinguish correct answers from errors
(Section~\ref{sec:confidence}). In the models we study, this mismatch grows
with scale. This suggests that the problem lies in how the model computes
its confidence, not in limited capacity, and that calibrating the output
confidence would not close the gap. Instead, the model's confidence would
need to track where the answer comes from, whether a stored fact, the
context, or neither.

%% file: sections/2_task.tex
\section{Experimental Design}
\label{sec:task}

\paragraph{Dataset.} We generated 1{,}600 associations of the form
\texttt{E000137} $\rightarrow$ \texttt{48210}. Identifiers are arbitrary
strings, and codes are five digits drawn uniformly at random. Nothing about
an entity predicts its code, so the model cannot infer these facts or
recover them from pretraining. It has either memorized a given mapping or it
has not, with nothing in between. We then ask
\emph{``What is \texttt{E000137}'s code?''} in four situations.

\begin{center}
\begin{tabular}{lcc}
\toprule
 & the fact is in the weights & the code is in the prompt \\
\midrule
\textbf{recall}             & yes & no \\
\textbf{reading}            & no  & yes \\
\textbf{knowledge conflict} & yes & yes, and it disagrees \\
\textbf{missing knowledge}  & no  & no \\
\bottomrule
\end{tabular}
\end{center}

\noindent In recall and reading, the correct answer is available to the model.
Knowledge conflict and missing knowledge are the two situations that lead to
confident errors, which we call \emph{interference} and
\emph{hallucination}, respectively. In knowledge conflict, the code in the
prompt is the stored code with its first digit changed. The two sources
therefore disagree at the first digit and agree at the remaining four.

\paragraph{Models.} We fine-tune Qwen2.5-3B-Instruct (36 layers, width 2048)
with rank-8 LoRA adapters for 10 epochs. Placing the adapter on QK, VO, MLP,
or all three (Full) installs the same knowledge through different parts of
the circuit. The adapters share every other hyperparameter, so any
difference in behavior comes from where the knowledge was installed (details
in Appendix~\ref{app:training}).

\paragraph{Training regimes.} We train the models in two ways. Under
\emph{single-template} training, every example uses the same phrasing,
\texttt{``Q: What is \{entity\}'s code? A:''}, repeated 100 times per entity.
This produces \emph{brittle} storage, in which recall is bound to the
training phrasing. Under \emph{varied-template} training, 53 phrasings span
direct questions, embedded statements, conversational turns, and distractor
prefixes. This produces \emph{robust} storage, in which recall survives
rephrasing. As we show in Section~\ref{sec:region}, this choice determines
whether a fact forms a basin at all, and it shapes most of our results.
Neither training set contains any example of conflict, refusal, or reading a
code from context, so the model is never explicitly taught how to behave in
those situations.

\paragraph{Evaluation.} We use greedy decoding and record \textbf{accuracy},
whether the generated string contains the target code. We also record the
error at each digit position and the rank of the correct token at the first
digit. Evaluation sets range from 30 to 1{,}600 queries depending on the
analysis (Appendix Table~\ref{tab:eval_sizes}). We measure the model's
\textbf{confidence} by the entropy of the next-token distribution over the
ten digit tokens at the first digit of the answer. This is what a monitor
that sees only the output would use. Section~\ref{sec:circuit} adds a third
measure, read from the hidden state rather than the output.

%% file: sections/3_circuit.tex
\section{Circuit Roles and Basin Geometry}
\label{sec:circuit}

To attribute an error to a specific component, we first need to know what
each component does during recall. In this section, we identify each
component's role in the pretrained model, following the circuits framework
\citep{elhage2021mathematical}. We then define basins and show that training
installs them.

\subsection{Jacobian Decomposition of Component Roles}
\label{sec:roles}
\label{sec:jacobian}

With the weights fixed, each sublayer applies a residual update
$h \mapsto h + f(h)$. This lets us view the transformer as a dynamical
system over depth \citep{geshkovski2023mathematical}. We extend this view to
generation, where the hidden state follows a trajectory under an
input-conditioned discrete-time system,
\begin{equation}
  h_{t+1} = F(h_t;\, x,\, y_{\le t}),
  \label{eq:dynamics}
\end{equation}
where $F$ composes the attention, feed-forward, residual, and normalization
maps. The local effect of a sublayer is given by its Jacobian $J$, which we
split into symmetric and antisymmetric parts,
\begin{equation}
  J = S + A, \qquad S = \tfrac{1}{2}(J + J^{\top}), \qquad A = \tfrac{1}{2}(J - J^{\top}).
  \label{eq:split}
\end{equation}
The symmetric part $S$ contracts or expands the hidden state, and the
antisymmetric part $A$ rotates it. We summarize each component by two
quantities. The first is its symmetry correlation
$\varphi = \mathrm{corr}(J_{ij}, J_{ji})$ over $i<j$, which ranges from $-1$
for a purely rotational map to $+1$ for a gradient-like one. The second is
$\|S\|_F^2$, the strength of its contraction.

\begin{figure}[t]
  \centering
  \includegraphics[width=0.9\textwidth]{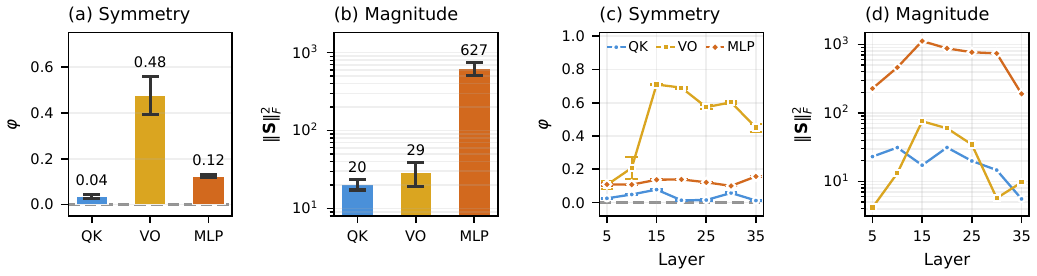}
  \caption{\textbf{Jacobian decomposition of the pretrained model.} Exact
    Jacobians at seven layers, averaged over five prompts. \emph{(a)}~Symmetry
    correlation $\varphi$. QK is unstructured, and VO is strongly symmetric.
    \emph{(b)}~Contracting magnitude $\|S\|_F^2$. MLP's is by far the largest. \emph{(c,d)}~The same quantities across layers. The three components thus
     have distinct dynamical signatures already in the pretrained model, MLP
     contracting toward stored states, QK steering, and VO copying content.}
  \label{fig:jacobian}
\end{figure}

Fig.~\ref{fig:jacobian} shows both quantities for the pretrained model,
computed from exact $2048\times2048$ Jacobians at seven layers. Each
component has a distinct signature, which we use to assign its role. MLP has
by far the largest symmetric part, at every layer
(Fig.~\ref{fig:jacobian}b,d). It is therefore the main force pulling the
hidden state toward particular regions, and we treat MLP as the main
substrate for \textbf{storing} facts. QK has near-zero symmetry and a small
symmetric part (Fig.~\ref{fig:jacobian}a). It does not reshape the hidden
state. Instead, it changes which token is attended to, and so
\textbf{steers} the trajectory toward stored content or toward the context.
VO is strongly symmetric, most of all in the middle layers
(Fig.~\ref{fig:jacobian}a,c). This is the signature of copying content along
fixed directions into the residual stream, so we treat VO as a
\textbf{readout} of retrieved content. This readout is distinct from the
output head, which converts the final hidden state into token probabilities
and sets the model's confidence.

These signatures appear in the pretrained model before any fine-tuning, and
the VO signature replicates across five model families
(Appendix~\ref{app:head_selection}). They therefore appear to be properties
of the pretrained architecture rather than artifacts of our fine-tuning,
while training determines what gets stored. Adapter placement supports this
assignment. Installing the same facts through MLP requires far fewer
gradient steps than through QK (Appendix~\ref{app:capacity}), and only
placements that include MLP produce recall that survives rephrasing
(Fig.~\ref{fig:dissociation}a). Placement shows what a component can learn
when it carries the task, however, so it constrains the component's role in
the pretrained model without uniquely determining it.

\subsection{Basins and the Geometric Margin}
\label{sec:region}

\begin{figure}[t]
  \centering
  \includegraphics[width=0.8\textwidth]{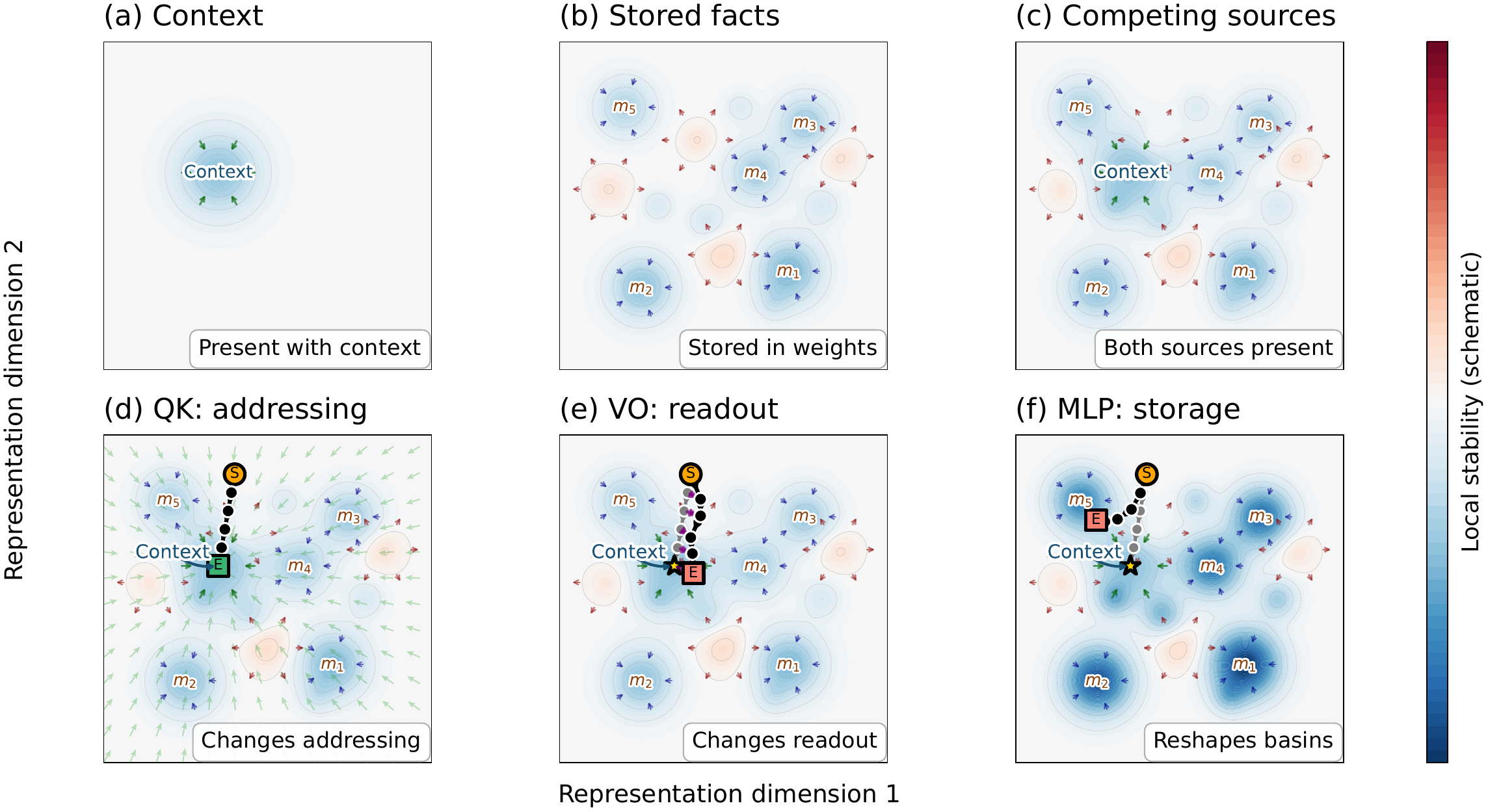}
  \caption{\textbf{Schematic basin geometry at a single layer.} The wells are
    basins in the finite-time sense of Section~\ref{sec:region}, regions the
    trajectory is pulled into at one step of generation, not a global landscape.
    \emph{(a)}~A code in context creates a transient basin, present only while
    those tokens are present. \emph{(b)}~Memorized facts create basins present
    for every input. \emph{(c)}~With both, the trajectory is pulled toward one
    or the other. \emph{(d)}~A QK adapter changes where the trajectory is
    steered, so it reaches the context. \emph{(e)}~A VO adapter changes what is
    read out, so the answer lands near, but not at, the target.
    \emph{(f)}~An MLP adapter deepens the memorized basins, which capture the
    trajectory before it reaches the context. Each placement thus fails in a
     different way, through where the trajectory is steered, what is read out,
     or how deep the stored basins are.}
  \label{fig:geometry}
\end{figure}

To ask whether the hidden state reached a stored fact, we need a reference
point for each fact. For each trained entity $i$, we average the final-layer
hidden state over three canonical phrasings and call the result its
\emph{memorized state} $m_i$. For a query $x$ with hidden state $h(x)$ at the
first-digit position, we define the \textbf{margin} as the distance to the
nearest memorized state, and the \textbf{gap} as how much farther the
second-nearest one is,
\begin{equation}
  \delta(x) = \min_i \|h(x) - m_i\|_2, \qquad
  \mathrm{gap}(x) = \delta_2(x) - \delta(x).
  \label{eq:margin}
\end{equation}
A small margin means that the query reached the vicinity of a memorized
representation, and a large margin means that it reached none. We call the
region around $m_i$ a fact's \emph{basin} (Fig.~\ref{fig:geometry}b). The
name comes from a basin of attraction, a region that hidden states are
pulled into and stay in. Three measurable properties justify the analogy. i) \emph{Contraction}: the MLP update pulls nearby
hidden states together, with most of its symmetric eigenvalues negative
through the middle layers. ii) \emph{Convergence}: as a result, hidden states
from a fact's many phrasings end up close to its memorized state.
Never-learned queries, by contrast, sit about as far from the nearest
memorized state as memorized states lie from one another
(Fig.~\ref{fig:metacognition}a). iii) \emph{Stability}: the robust store
keeps producing its answer under more embedding noise than the brittle store
(Appendix Fig.~\ref{fig:perturbation_stability}). Unlike a basin of
attraction, however, a basin here exists only over a finite number of layers
and generation steps, since the composed map has no fixed points
(Appendix~\ref{app:basin}). The context can create a basin as well. A code
given in the prompt pulls the hidden state toward it while those tokens are
present (Fig.~\ref{fig:geometry}a). Under knowledge conflict, the trajectory
is therefore pulled toward two basins at once (Fig.~\ref{fig:geometry}c).

Two controls show that basins are installed by training on varied phrasings,
rather than read off the entity strings. The margin does not separate
recalled from never-learned queries in the base model. Nor does it separate
them in brittle adapters, even though these recall their own phrasing
perfectly (Appendix~\ref{app:basin}).

We use these roles to interpret the errors in Section~\ref{sec:failures}.
Fig.~\ref{fig:geometry}d--f summarizes what an adapter on each component
changes. A QK adapter changes which basin the trajectory reaches, a VO
adapter changes what is read out from it, and an MLP adapter reshapes the
basins themselves.

%% file: sections/4_failures.tex
\section{Failure Modes}
\label{sec:failures}

In this section, we examine the two situations that lead to confident
errors. In both, the component roles of Section~\ref{sec:circuit} determine
how the model fails. Under knowledge conflict (Section~\ref{sec:conflict}),
the two sources interfere. How robustly a fact is stored decides whether the
weights or the context win, and each adapter placement leaves its own error
pattern. Under missing knowledge (Section~\ref{sec:hallucination}), the
hidden state reaches no basin, yet the output head still commits to an
answer. The placement determines how confident that answer is.

\subsection{Knowledge Conflict and Interference}
\label{sec:conflict}

\begin{figure}[t]
  \centering
  \includegraphics[width=0.9\textwidth]{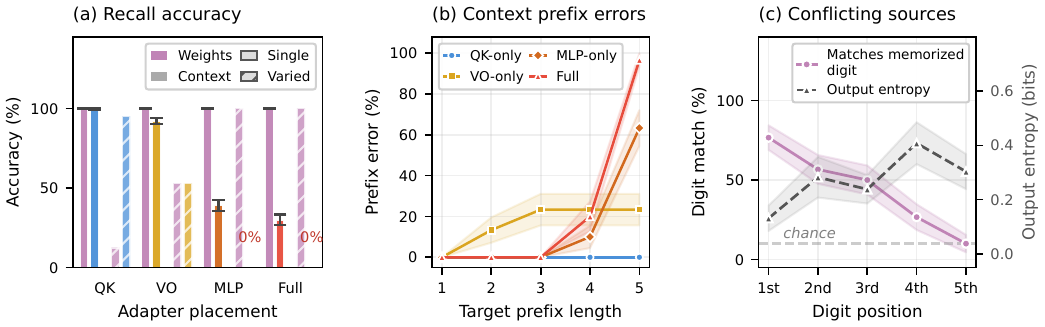}
  \caption{\textbf{Interference by adapter placement under knowledge conflict.} \emph{(a)}~Recall from the weights and from the
    context by placement, under single-template (brittle) and varied-template
    (robust) training. QK preserves context reading but cannot store facts
    robustly, while storing facts robustly through MLP or Full drives context
    reading to zero. \emph{(b)}~Prefix error, the fraction of answers whose first
    $k$ digits do not match the code in the context, for brittle adapters
    (30 queries each). QK makes no errors, VO's errors appear by the third
    digit and do not grow further, and MLP and Full make late errors that
    compound. \emph{(c)}~Under a contradiction with a robustly
    stored fact, the answer begins with the stored code and drifts away from
    it, while output entropy stays low. The form of interference is therefore
     set by where and how robustly the fact is stored, and entropy does not
     reveal it.}
  \label{fig:dissociation}
\end{figure}

Under knowledge conflict, the correct answer is always stored in the weights.
What fails is the arbitration between the weights and the context. The
resulting interference takes one of three forms, depending on how robustly
the fact is stored (Fig.~\ref{fig:dissociation}). The stored fact can
\emph{corrupt} the reading of the context, \emph{override} the context, or
make the answer \emph{drift} into one that matches neither source.

\paragraph{Brittle storage.} When a fact is learned from a single phrasing,
the context wins, but the placements that reshape the basins corrupt the end
of the answer. Brittle adapters recall stored facts on the training phrasing
but almost never under any other phrasing. Their basins are therefore
inactive in a new prompt, and in a direct contradiction the stored code never
wins (Appendix Table~\ref{tab:conflict_instructions}). The placement instead
determines how well the model reads the context
(Fig.~\ref{fig:dissociation}a). QK reads it almost perfectly and VO nearly as
well, while MLP and Full read it much less reliably. The placement also
determines where the errors appear (Fig.~\ref{fig:dissociation}b), and each
pattern matches the role of the component (Fig.~\ref{fig:geometry}d--f). QK,
which only steers, makes almost no errors. VO, which reads out content, makes
errors that appear early and stop growing, consistent with a readout that is
slightly imprecise at every step. MLP and Full, which reshape the basins,
read the first digits correctly but fail at the last ones. Once enough digits
have been generated, the stored codes appear to capture the trajectory
(sample outputs in Appendix~\ref{app:samples}).

\paragraph{Robust storage.} When a fact is learned from many phrasings, the
stored fact overrides the context. MLP and Full recall every stored fact
across phrasings but no longer read the context at all, which we call
\emph{context deafness} (Fig.~\ref{fig:dissociation}a). QK shows the
opposite tradeoff. It reads the context reliably but recalls few stored
facts, since steering alone does not create basins. VO falls in between. The
component that reshapes the basins, and therefore stores facts robustly, is
thus the same one that blocks the context.

\paragraph{Contradiction under robust storage.} When a robustly stored fact
meets a contradicting context, the answer starts on the stored code and then
drifts into one that matches neither source (Fig.~\ref{fig:dissociation}c).
The code in the prompt differs from the stored code only in its first digit,
so the two sources disagree at the first digit and agree at the rest. The
model usually follows its weights at the first digit. At the remaining
digits, however, where both sources agree, it increasingly fails to
reproduce the stored code. Almost no answer matches either source in full,
and the entropy stays low throughout. Because this error unfolds over the
answer, it is not visible at the first digit. There, the query is still
nearest its own memorized state, and signals read at that point separate
these errors poorly (Appendix~\ref{app:layer_heatmap}).

We conclude that interference is a failure of arbitration rather than of
storage. The outcome of this arbitration is set by how the fact is stored,
not by what the two sources say. Retrieval fails even at the digits where the
two sources agree. Under brittle storage, giving the same code in both does
not reliably help and can even hurt (Appendix~\ref{app:agreement}). The
generation trajectory is therefore largely fixed by how the fact is stored,
while the entropy stays low.

\subsection{Missing Knowledge and Hallucination}
\label{sec:hallucination}

\begin{figure}[t]
  \centering
  \includegraphics[width=0.9\textwidth]{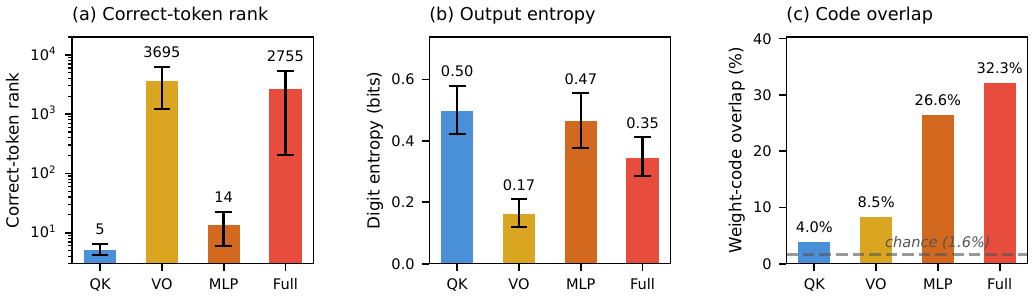}
  \caption{\textbf{Hallucination by adapter placement.} Never-learned entities.
    \emph{(a)}~Rank of the correct token. VO and Full push it far down the ranking, while QK and MLP keep it near the top.
    \emph{(b)}~Digit entropy. VO is the most confident, and QK and MLP the least. \emph{(c)}~Fraction of hallucinated codes that
    match a memorized code. MLP and Full reproduce memorized codes far more often than chance. Every placement hallucinates, but VO does
     so most confidently and with the correct answer least accessible.}
  \label{fig:hallucination}
\end{figure}

Under missing knowledge, there is no stored fact to retrieve. The hidden
state reaches no basin, yet the output head still commits to an answer, and
the adapter placement determines how confident that answer is
(Fig.~\ref{fig:hallucination}).

\paragraph{Basin absence.} Without an adapter, the base model declines to
answer almost all queries about entities it has never seen
(Appendix~\ref{app:samples}). With any adapter, it instead produces a
confident five-digit code (Appendix Table~\ref{tab:samples_hallucination}),
even though the hidden state never reaches a basin. For the robust MLP
adapter, never-learned queries lie about as far from the nearest memorized
state as memorized states lie from one another, and none falls within the
range of recalled queries. A directional analysis suggests why the answer is
confident anyway. The output head is sensitive only to particular
projections of the hidden state. When a hallucination reproduces another
entity's code, its hidden state is far from that entity's memorized state.
Along the direction that selects the emitted digit, however, the two states
nearly agree, and both assign that digit a high logit. Token-level
confidence therefore cannot tell reaching a basin apart from merely
reproducing its code.

\paragraph{Effect of adapter placement.} Every placement hallucinates, but
the placement determines the character of the hallucination
(Fig.~\ref{fig:hallucination}). VO, which writes content into the residual
stream, produces the most confident hallucinations and pushes the correct
token far down the ranking. QK and MLP are the least confident and keep the
correct token near the top. MLP and Full, which reshape the basins globally,
bias the answers toward memorized codes and reproduce one far more often than
chance. This bias arises at the output rather than from capture by a
specific basin, since the reproduced code rarely belongs to the nearest
stored entity.

We conclude that, in our setting, the model's output does not reliably
distinguish a learned fact from a never-learned one. Once any component is
fine-tuned, the model stops declining and answers never-learned queries just
as it answers learned ones. Its output head commits to an answer whether or
not the hidden state reached a basin. The model behaves as if it had no step
that checks whether the fact is stored, and Section~\ref{sec:confidence}
examines this metacognitive gap.

%% file: sections/5_confidence.tex
\section{Why Output Confidence Fails}
\label{sec:confidence}

Section~\ref{sec:failures} shows that the model can stay confident in both
interference and hallucination. In this section, we first ask why the output
entropy does not flag these errors (Section~\ref{sec:pathway}). We then ask
what can flag them instead (Section~\ref{sec:helps}) and how the problem
changes with scale (Section~\ref{sec:scale}).

\subsection{Why Entropy Misses the Errors}
\label{sec:pathway}

To understand why the entropy fails to flag these errors, we first test
whether the information that separates a correct answer from an error is
lost before the output. We train linear probes to classify correct versus
incorrect answers across all five situations, one from the hidden state and
one from the output logits. Over all answers, both probes and the entropy
perform almost equally well. The difference appears on the answers given with
high confidence. These are selected for low entropy, so the entropy
separates them poorly by construction, but both probes still separate them
well. The model therefore encodes whether a confident answer is correct, even
though the entropy does not reflect it.

The reason is that entropy only measures how peaked the distribution over the
ten digits is. This depends on how strongly the final hidden state points
toward one digit, not on whether it reached a basin. As
Section~\ref{sec:hallucination} shows, the directions that encode whether a
fact was stored barely affect the digit logits. Catching every hallucination
by entropy would therefore also reject nearly all correct recalls
(Fig.~\ref{fig:metacognition}b). The same reasoning applies to interference.
As the answer drifts, the output head commits to each digit as sharply as
under clean recall (Fig.~\ref{fig:dissociation}c), because interference
changes which digit is chosen, not how sharply.

\subsection{Detecting the Errors}
\label{sec:helps}

Section~\ref{sec:pathway} suggests that a better detector should read whether
the hidden state reached a basin, rather than how peaked the output is. We
compare three ways to obtain such a signal: i) the margin, ii) a head trained
to flag errors, and iii) consistency across paraphrases. i) The margin
separates recalled from never-learned queries almost perfectly
(Fig.~\ref{fig:metacognition}a,b). It can do so only because it is given the
memorized state of every stored fact. Without a fact's own memorized state,
it falls to chance, so we use the margin to study the geometry rather than as
a practical detector. ii) We also train a head on the hidden states to
predict whether an answer is correct, jointly with the adapter. This head
does not do better than the entropy (Appendix~\ref{app:e2e_head}), which
suggests that correctness labels alone do not teach the model to read whether
a basin was reached. iii) Consistency across paraphrases needs only the
model's outputs. Every phrasing of a stored fact converges into the same
basin, so the model should give the same answer however the question is
phrased. A never-learned entity has no basin to converge to, so different
phrasings should end in different places. We therefore ask the same question
in eight phrasings and count the distinct answers. For stored facts, the
model gives the same code in nearly all phrasings, while for never-learned
entities the answers vary much more. The count separates the two well.
Unlike the margin, it needs no list of stored facts, and it generalizes to
new facts. It is a paraphrase-based variant of self-consistency and semantic
entropy \citep{wang2022self,kuhn2023semantic}.

\subsection{Scaling of Confident Hallucination}
\label{sec:scale}

Finally, we ask whether larger models fix this problem. We could not reliably
measure basins in natural-language models (Appendix~\ref{app:nl-withdrawn}),
so we instead test one prediction of our account on real models, using the
full next-token entropy at the first answer token. The benchmarks are
closed-book, so no context can conflict with the weights, and we count every
wrong answer as a hallucination. Our account predicts that scale makes the
problem worse rather than better. In the models we evaluate, larger models
separate their top two logits by a wider gap (Appendix~\ref{app:scaling}).
The softmax turns a wider gap into a sharper distribution whether or not the
answer sits in a basin. Larger models should therefore deliver a larger
fraction of their hallucinations with high confidence.

\begin{figure}[t]
  \centering
  \includegraphics[width=0.9\textwidth]{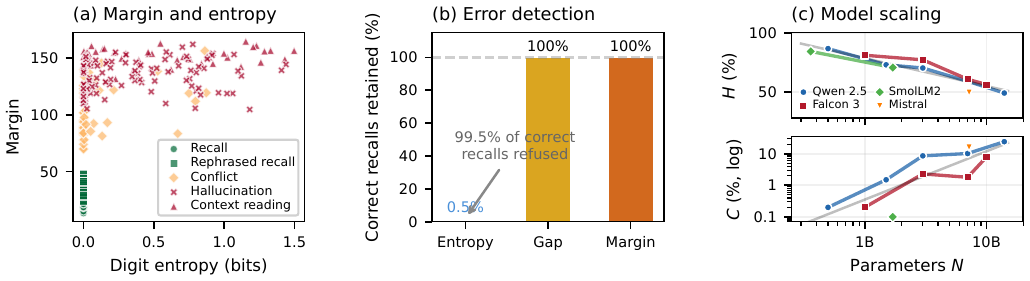}
  \caption{\textbf{Margin versus entropy, and confident hallucination across scale.}
    \emph{(a)}~Recall clusters at low margin and hallucination at high margin.
    \emph{(b)}~Catching every hallucination by entropy rejects nearly all correct recalls, while the margin rejects none. \emph{(c)}~Across 12 models, hallucination
    rate $H$ falls while confident fraction $C$ rises. The geometric margin
     separates hallucination from recall where entropy cannot, and as models
     scale, the hallucinations that remain are increasingly confident.}
  \label{fig:metacognition}
\end{figure}

To make this prediction quantitative, let $\bar{\Delta}$ be the mean gap
between the top two logits on wrong answers, and let $\Delta^*$ be the gap at
which the entropy falls below a threshold $H_0$. The fraction of confident
hallucinations is then the fraction of wrong answers whose gap exceeds
$\Delta^*$, that is, $C = \Pr[\Delta > \Delta^*]$. If these gaps are
exponentially distributed, this gives
\begin{equation}
  C = \exp\!\left(-\frac{c}{\bar{\Delta}}\right), \qquad c = \Delta^*(H_0, V),
  \label{eq:law}
\end{equation}
with no free parameter once $\Delta^*$ is measured. The measured gaps are
close to exponential but not exactly so (Appendix~\ref{app:scaling}), so we
expect Eq.~\ref{eq:law} to capture the trend rather than exact values.
Fig.~\ref{fig:metacognition}c shows that it does. Across 12
instruction-tuned models from four families and two benchmarks, the points
follow Eq.~\ref{eq:law} closely.

In a teacher--student ablation where only the width changes, the margin
separates never-learned from recalled queries more and more sharply as the
width grows, while $C$ approaches one (Appendix~\ref{app:scaling}). Since
width alone produces this divergence, the rise in confident hallucination
across real models need not come from differences in their data or training
recipes. At least in this controlled setting, the information needed to
detect hallucinations also becomes more available with scale, even as the
entropy becomes less informative. Larger models do hallucinate less overall,
but only slowly (Appendix~\ref{app:scaling}). Since $C$ rises faster, the
fraction of all questions answered with a confident hallucination,
$H \cdot C$, grows with scale. The limitation therefore lies in how
confidence is read out, not in what the model represents, and in the models
we study it becomes more costly with scale.

%% file: sections/6_related.tex
\section{Related Work}
\label{sec:related}

Prior work shows that models can be trained to predict whether they know an
answer \citep{kadavath2022language}. Other studies find that internal states
encode correctness better than output probabilities
\citep{azaria2023internal,orgad2025}, and that fine-tuning on new knowledge
increases hallucination \citep{gekhman2024fine}. These studies work in
natural language, where what the model has learned is uncertain. We instead
isolate this gap with exact ground truth and localize it to specific circuit
components. Our storage claim builds on key--value accounts of feed-forward
layers \citep{geva-etal-2021-transformer,meng2022locating,meng2022mass}. Our
interference results relate to work on knowledge conflict, which asks when
the weights or the context win
\citep{longpre2021entity,neeman2023disentqa,li2022large,xie2024adaptive}.
Calibration methods \citep{guo2017calibration} inherit the blind spot of the
output head, whereas hidden-state probes \citep{slobodkin2023,kossen2024semantic}
and self-consistency \citep{wang2022self,kuhn2023semantic} do not.
Appendix~\ref{app:related} discusses related work in more detail.

%% file: sections/7_discussion.tex
\section{Discussion}
\label{sec:discussion}

\paragraph{Key takeaway.} In both interference and hallucination, the model
encodes information that separates its correct answers from its errors, but
its confidence does not use it. Under missing knowledge, the hidden state
lies outside every basin. In both situations, probes on the hidden state and
on the output logits separate correct answers from errors, even among the
answers given with high confidence. The entropy, however, only measures how
peaked the output distribution is, and a contradicted or never-learned query
can produce a distribution as peaked as a known one. Since outputs sharpen
with scale whether or not a fact was retrieved, a growing fraction of the
hallucinations of larger models are confident.

\paragraph{Implications.} Our findings have two practical implications, one
for how knowledge is added to a model and one for how its errors are
detected. i) For knowledge injection, where and how fine-tuning installs a
fact determines how the model later arbitrates between its weights and its
context. Robust MLP storage can make the model ignore context meant to
correct it, as in retrieval-augmented generation. Moreover, every adapter
placement made the model answer questions that the base model declined.
Evaluations of knowledge injection should therefore test context reading and
unknown queries, not only recall. ii) For hallucination detection, signals
that read whether a fact was reached, such as the margin or paraphrase
consistency, are a more promising basis than output confidence, which also
becomes less informative as models grow. Interference, by contrast, unfolds
during generation, as when the answer drifts into one that matches neither
source. Detecting it therefore likely requires monitoring every generated
token rather than only the first.

\paragraph{Toward metacognition.} For a model to know what it knows, its
confidence should be computed from whether its hidden state reached stored
content, not from how peaked its output is. Post-hoc calibration cannot
provide this, since it does not change which answers look confident, and our
scaling results suggest that size alone will not either. We see three
directions. i) Architectures could expose a retrieval signal explicitly. For
example, memory components could report how well a query matches stored
content, so that the output head can condition on whether a basin was
reached. ii) Training could target this signal directly. A head trained end
to end on correctness labels did not surpass the entropy in our setting
(Appendix~\ref{app:e2e_head}), which suggests that the signal should be
supervised at the level of retrieval. One option is to pair queries about
unseen entities with abstention targets during fine-tuning, so that
knowledge injection does not erase the base model's ability to decline.
iii) The answers to a stored fact agree across phrasings, while those to a
never-learned entity do not. A model could therefore estimate its confidence
from how well its own trajectories agree. Under knowledge conflict, it could
also track which source it follows at every generated token, rather than
only at the first. Our experiments use synthetic facts, so that we know
exactly what the model has learned. Testing these directions on naturally
learned facts is the main next step. It will require a way to determine what
a pretrained model knows that does not depend on how the question is worded
(Appendix~\ref{app:nl-withdrawn}).

\section{Conclusion}
\label{sec:conclusion}

In our experiments, we controlled what a language model knows, how robustly
it knows it, and where in the network the knowledge is stored. We used this
control to follow how the model answers when its context contradicts a
stored fact and when it never learned the fact at all. In both cases, the
model's hidden states and output logits carry information that separates
correct answers from errors. Its confidence, measured by the entropy of the
output distribution, reflects only how sharply it commits to an answer, so it
can be as confident when it is wrong as when it is right. In the models we
study, this gap between confidence and correctness widens with scale.
Methods that recalibrate the output after training, for example with temperature
scaling, do not close the metacognitive gap, since recalibration changes the
confidence values without changing which answers look confident. To know what
it knows, a language model needs a confidence that tracks where its answers
come from. We show that information separating correct answers from errors
is partially available in the model's hidden states, which suggests that a
better confidence signal can be derived from what the model already
computes. The challenge ahead is how to use this information when the model
decides how confident to be.

%% file: appendix.tex
\appendix

\section{Training Details}
\label{app:training}

\subsection{Prompt templates for robust storage}
\label{app:templates}

Robust storage training uses 53 prompt templates organized into six categories, designed to cover a wide range of natural prompt structures. Each training example randomly samples one template; additionally, 25\% of examples undergo synonym paraphrasing (substituting ``code'' $\to$ ``identifier''/``number''/``value''/``designation'', etc.) and 10\% receive a distractor prefix (e.g., ``Note: The system is operational.''). Table~\ref{tab:templates} lists all 53 templates.

\begin{table}[h]
\centering
\scriptsize
\begin{tabular}{p{0.15\textwidth} p{0.78\textwidth}}
\toprule
\textbf{Category} & \textbf{Templates} \\
\midrule
\textbf{A. Direct queries} (18) & \texttt{Q: What is \{entity\}'s code? A:} \enspace$\cdot$\enspace \texttt{Q: What is the code for \{entity\}? A:} \enspace$\cdot$\enspace \texttt{What is \{entity\}'s code?} \enspace$\cdot$\enspace \texttt{What is the code for \{entity\}?} \enspace$\cdot$\enspace \texttt{Entity \{entity\}'s code is?} \enspace$\cdot$\enspace \texttt{Give the code associated with \{entity\}.} \enspace$\cdot$\enspace \texttt{Return the code for entity \{entity\}.} \enspace$\cdot$\enspace \texttt{Provide the code for \{entity\}.} \enspace$\cdot$\enspace \texttt{Tell me \{entity\}'s code.} \enspace$\cdot$\enspace \texttt{What code corresponds to \{entity\}?} \enspace$\cdot$\enspace \texttt{The code for \{entity\} is:} \enspace$\cdot$\enspace \texttt{\{entity\} maps to what code?} \enspace$\cdot$\enspace \texttt{Which code is assigned to \{entity\}?} \enspace$\cdot$\enspace \texttt{What is the identifier for \{entity\}?} \enspace$\cdot$\enspace \texttt{What is the mapping for \{entity\}?} \enspace$\cdot$\enspace \texttt{What is the designation for \{entity\}?} \enspace$\cdot$\enspace \texttt{\{entity\}'s identifier:} \enspace$\cdot$\enspace \texttt{\{entity\}'s mapping:} \\
\addlinespace
\textbf{B. Embedded statements} (10) & \texttt{Given the list below, what is the code for \{entity\}?} \enspace$\cdot$\enspace \texttt{The records show: \{entity\} → ?} \enspace$\cdot$\enspace \texttt{Find the appropriate code for \{entity\} from the database.} \enspace$\cdot$\enspace \texttt{Look up the code for \{entity\} in the system.} \enspace$\cdot$\enspace \texttt{According to our records, \{entity\}'s code is:} \enspace$\cdot$\enspace \texttt{In the registry, \{entity\} corresponds to:} \enspace$\cdot$\enspace \texttt{The system maps \{entity\} to:} \enspace$\cdot$\enspace \texttt{Database entry for \{entity\}:} \enspace$\cdot$\enspace \texttt{Retrieve the code associated with \{entity\}.} \enspace$\cdot$\enspace \texttt{Query result for \{entity\}:} \\
\addlinespace
\textbf{C. Conversational} (7) & \texttt{User: I need the code for \{entity\}. Assistant:} \enspace$\cdot$\enspace \texttt{User: What's \{entity\}'s code? Assistant:} \enspace$\cdot$\enspace \texttt{User: Can you look up \{entity\} for me? Assistant:} \enspace$\cdot$\enspace \texttt{User: Tell me about \{entity\}. Assistant: \{entity\}'s code is} \enspace$\cdot$\enspace \texttt{In our system, how would you encode \{entity\}?} \enspace$\cdot$\enspace \texttt{If I asked for \{entity\}, what code would you give me?} \enspace$\cdot$\enspace \texttt{Someone is asking about \{entity\}. What code should I tell them?} \\
\addlinespace
\textbf{D. Instructional} (6) & \texttt{Provide the code that corresponds to \{entity\}.} \enspace$\cdot$\enspace \texttt{List the code for \{entity\}.} \enspace$\cdot$\enspace \texttt{Output \{entity\}'s code.} \enspace$\cdot$\enspace \texttt{Display the code for \{entity\}.} \enspace$\cdot$\enspace \texttt{Show me \{entity\}'s code.} \enspace$\cdot$\enspace \texttt{Print the code for \{entity\}.} \\
\addlinespace
\textbf{E. Completion-style} (5) & \texttt{\{entity\}'s code is} \enspace$\cdot$\enspace \texttt{\{entity\} →} \enspace$\cdot$\enspace \texttt{\{entity\} maps to} \enspace$\cdot$\enspace \texttt{Code for \{entity\}:} \enspace$\cdot$\enspace \texttt{The code is} \\
\addlinespace
\textbf{F. Distractor-laden} (7) & \texttt{The weather is nice today. What is \{entity\}'s code?} \enspace$\cdot$\enspace \texttt{Bob read about something yesterday. Also, the code for \{entity\} is:} \enspace$\cdot$\enspace \texttt{Alice owns a car. The code for \{entity\} is:} \enspace$\cdot$\enspace \texttt{Random note: the sky is blue. Now, what is the code for \{entity\}?} \enspace$\cdot$\enspace \texttt{First, note that 2+2=4. Second, \{entity\}'s code is:} \enspace$\cdot$\enspace \texttt{Consider the following: \{entity\} has the code:} \enspace$\cdot$\enspace \texttt{Unrelated: cats are animals. Relevant: \{entity\}'s code?} \\
\bottomrule
\end{tabular}
\caption{All 53 prompt templates used for robust storage training. Each training example randomly samples one template and optionally applies synonym paraphrasing (25\% probability) and distractor prefix injection (10\% probability). The training data contains zero conflict examples, zero refusal examples, and zero context-reading examples.}
\label{tab:templates}
\end{table}

\subsection{Data generation}
\label{app:data}

\paragraph{Entity--code assignments.}
Each entity is assigned an identifier of the form \texttt{E\{i:06d\}} (e.g., \texttt{E000000}--\texttt{E001599}) and a random 5-digit code sampled uniformly from $\{10000, \dots, 99999\}$. With $N = 1{,}600$ entities, we obtain 1{,}586 unique codes out of 90{,}000 possible strings (1.76\% coverage), so chance overlap between codes is low.

\paragraph{Brittle storage data.}
A single template (``\texttt{Q: What is \{entity\}'s code? A:}'') is used for all training examples. Each entity is repeated 100 times, yielding $N \times 100$ training examples (e.g., 160{,}000 for $N = 1{,}600$). Examples are shuffled before training.

\paragraph{Robust storage data.}
The same $N = 1{,}600$ entity--code mappings are used with all 53 templates. Each of the 100 repetitions per entity randomly samples a template from the full set. With synonym paraphrasing (25\%) and distractor prefixes (10\%), this yields 160{,}000 examples, split 95/5 into 152{,}000 training and 8{,}000 validation examples.

\paragraph{Held-out evaluation sets.}
Five evaluation sets are constructed, each with 50 samples: (i) format generalization (novel templates never seen in training), (ii) distractor robustness, (iii) knowledge conflict, (iv) unknown/unseen entities, and (v) paraphrase variants.

\subsection{LoRA hyperparameters}
\label{app:hyperparams}

Table~\ref{tab:hyperparams} summarizes the LoRA configuration for all experiments. All adapters use the same optimizer settings; only target modules and rank differ.

\begin{table}[h]
\centering
\small
\begin{tabular}{ll}
\toprule
\textbf{Parameter} & \textbf{Value} \\
\midrule
Base model & Qwen2.5-3B-Instruct ($d = 2{,}048$, 36 layers) \\
LoRA rank $r$ & 8 (main experiments); 32, 64 (rank sweep) \\
LoRA $\alpha$ & $2r$ (i.e., 16, 64, 128) \\
LoRA dropout & 0.0 \\
Learning rate & $5 \times 10^{-5}$ \\
LR scheduler & Cosine with 10\% warmup \\
Max gradient norm & 1.0 \\
Batch size & 16 per device \\
Epochs & 10 \\
Max sequence length & 64 tokens \\
Optimizer & AdamW (HuggingFace default) \\
Precision & bf16 \\
Training loss & Completion-only (prompt tokens masked) \\
\bottomrule
\end{tabular}
\caption{LoRA hyperparameters shared across all adapter types. Target modules vary by adapter type: QK = \{\texttt{q\_proj}, \texttt{k\_proj}\}; VO = \{\texttt{v\_proj}\}; MLP = \{\texttt{gate\_proj}, \texttt{up\_proj}, \texttt{down\_proj}\}; Full = all of the above.}
\label{tab:hyperparams}
\end{table}

\subsection{Evaluation protocol}
\label{app:eval}

All evaluations use greedy decoding (\texttt{max\_new\_tokens} $= 8$). We report two accuracy metrics:
\begin{itemize}
\item \textbf{Exact match:} The generated string exactly equals the target code.
\item \textbf{Contains:} The target code appears as a substring of the generated response (more lenient, accounts for minor formatting differences).
\end{itemize}

For knowledge conflict evaluation, we additionally classify each response as:
\begin{itemize}
\item \textbf{Context wins:} The response contains the context-provided code $c_{\mathrm{ctx}}$.
\item \textbf{Weights win:} The response contains the parametrically memorized code $c_{\mathrm{stored}}$.
\item \textbf{Neither:} The response contains neither code (garbage output or unrelated text).
\end{itemize}

\paragraph{Evaluation set sizes.} Table~\ref{tab:eval_sizes} lists the number
of queries behind each analysis.

\begin{table}[h]
\centering
\small
\begin{tabular}{p{4.1cm}p{2.0cm}p{3.5cm}p{2.5cm}}
\toprule
Analysis & Adapters & Situations & Queries \\
\midrule
Recall and reading accuracy (Fig.~\ref{fig:dissociation}a) & brittle & recall / reading / conflict / missing & 1{,}600 / 200 / 200 / 200 \\
Per-digit and mechanistic (Figs.~\ref{fig:dissociation}b, \ref{fig:hallucination}; App.~\ref{app:agreement}) & brittle & each situation & 30 \\
Robust behavioral sets (Fig.~\ref{fig:dissociation}a) & robust & five held-out sets & 50 each \\
Robust conflict per digit (Fig.~\ref{fig:dissociation}c) & robust MLP & conflict & 30 \\
Geometry and probes (Secs.~\ref{sec:region}, \ref{sec:confidence}; Fig.~\ref{fig:metacognition}a,b) & robust MLP & recall / novel phrasing / conflict / missing / reading & 200 / 50 / 50 / 100 / 50 \\
Scaling relation (Sec.~\ref{sec:scale}; Fig.~\ref{fig:metacognition}c) & 12 pretrained models & TriviaQA, NQ-Open & 1{,}500 each \\
Hallucination rate (App.~\ref{app:scaling}) & 17 pretrained models & TriviaQA & 2{,}000 \\
\bottomrule
\end{tabular}
\caption{Number of queries behind each analysis.}
\label{tab:eval_sizes}
\end{table}

\section{Capacity Scaling Experiments}
\label{app:capacity}

To understand how memorization efficiency varies with adapter architecture and rank, we conduct systematic scaling experiments. All adapters achieve near-perfect final training loss (${\approx}0.0001$) across all tested $N$; the meaningful difference is \emph{learning efficiency}, that is, how many gradient steps are required to memorize $N$ entities.

\subsection{Module ablation: scaling entities at rank 8}
\label{app:capacity_modules}

\begin{figure}[h]
  \centering
  \includegraphics[width=\textwidth]{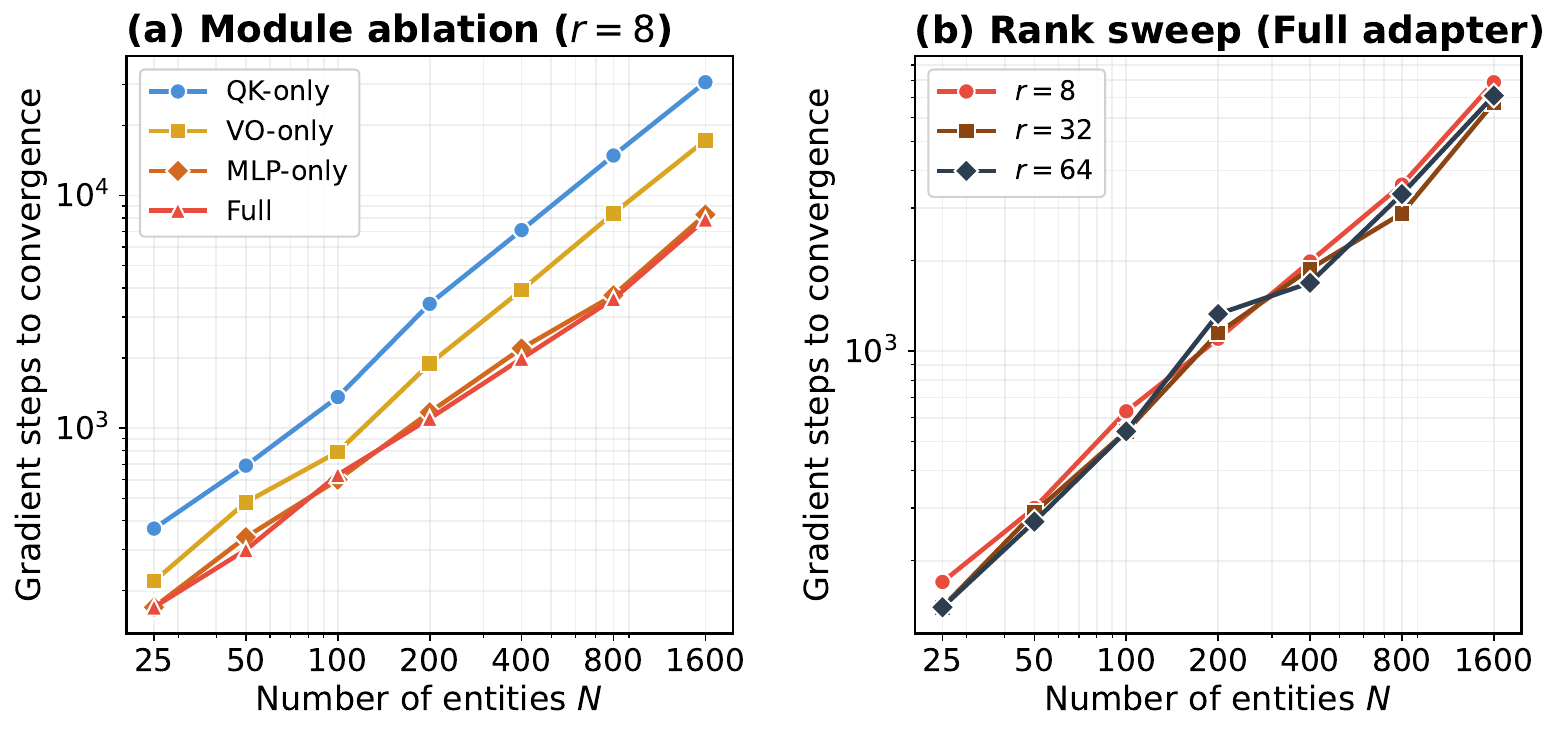}
  \caption{\textbf{Learning efficiency scaling curves.} Gradient steps to first reach training loss $< 0.05$ (log--log scale). All adapters achieve near-zero final loss for all $N$; the y-axis captures how efficiently each adapter memorizes.
  \emph{(a)}~Adapter placement at $r = 8$. MLP and Full require far fewer steps than QK, consistent with MLP being the main substrate for storing associations. Steps scale approximately linearly with $N$ for all adapters.
  \emph{(b)}~Rank sweep for the Full adapter. Rank has little effect on learning efficiency, with at most ${\sim}15\%$ fewer steps at higher rank. Where the adapter is placed, not its capacity, determines how efficiently facts are stored, and MLP is the most efficient placement.}
  \label{fig:capacity_scaling}
\end{figure}

We train all four adapter types (QK, VO, MLP, Full) at rank $r = 8$ across seven entity counts: $N \in \{25, 50, 100, 200, 400, 800, 1{,}600\}$, each with 100 repetitions per entity. Table~\ref{tab:capacity_loss} reports steps to convergence (first step where training loss $< 0.05$) and Figure~\ref{fig:capacity_scaling}a visualizes the scaling curves.

\begin{table}[h]
\centering
\small
\begin{tabular}{lrrrrrrr}
\toprule
& \multicolumn{7}{c}{Number of entities $N$} \\
\cmidrule(lr){2-8}
Adapter & 25 & 50 & 100 & 200 & 400 & 800 & 1{,}600 \\
\midrule
QK  &   370 &   690 & 1{,}360 & 3{,}420 & 7{,}090 & 14{,}830 & 30{,}620 \\
VO   &   220 &   480 &   790 & 1{,}890 & 3{,}920 &  8{,}370 & 17{,}240 \\
MLP &   170 &   340 &   600 & 1{,}170 & 2{,}200 &  3{,}720 &  8{,}260 \\
Full     &   170 &   300 &   630 & 1{,}100 & 1{,}990 &  3{,}590 &  7{,}880 \\
\bottomrule
\end{tabular}
\caption{Gradient steps to convergence (first step with training loss $< 0.05$) by adapter type and entity count ($r = 8$, 10 epochs). All adapters reach near-zero final loss; the table captures learning efficiency. MLP and Full require far fewer steps than QK (8{,}260 and 7{,}880 versus 30{,}620 at $N = 1{,}600$), consistent with MLP being the main substrate for storing associations.}
\label{tab:capacity_loss}
\end{table}

\section{Conflict Arbitration Details}
\label{app:conflict}

\subsection{Instruction sensitivity}

We test whether explicit instructions can influence conflict resolution under brittle storage. Table~\ref{tab:conflict_instructions} shows how often the context wins across three instruction conditions.

\begin{table}[h]
\centering
\small
\begin{tabular}{lrrr}
\toprule
Adapter & Default & ``Use notes'' & ``Use own knowledge'' \\
\midrule
QK  & 100\% & 93\% & 100\% \\
VO   & 93\%  & 97\% & 97\%  \\
MLP & 50\%  & 40\% & 50\%  \\
Full     & 33\%  & 40\% & 43\%  \\
\bottomrule
\end{tabular}
\caption{Rate at which the context wins under conflict with varying instructions ($N = 30$ per condition, brittle storage). Instructions have minimal effect, and Weights win 0\% across all 360 trials. For MLP and Full, the ``neither'' category (garbage output) accounts for the remainder.}
\label{tab:conflict_instructions}
\end{table}

\subsection{Agreement between weights and context}
\label{app:agreement}

\begin{figure}[h]
  \centering
  \includegraphics[width=\textwidth]{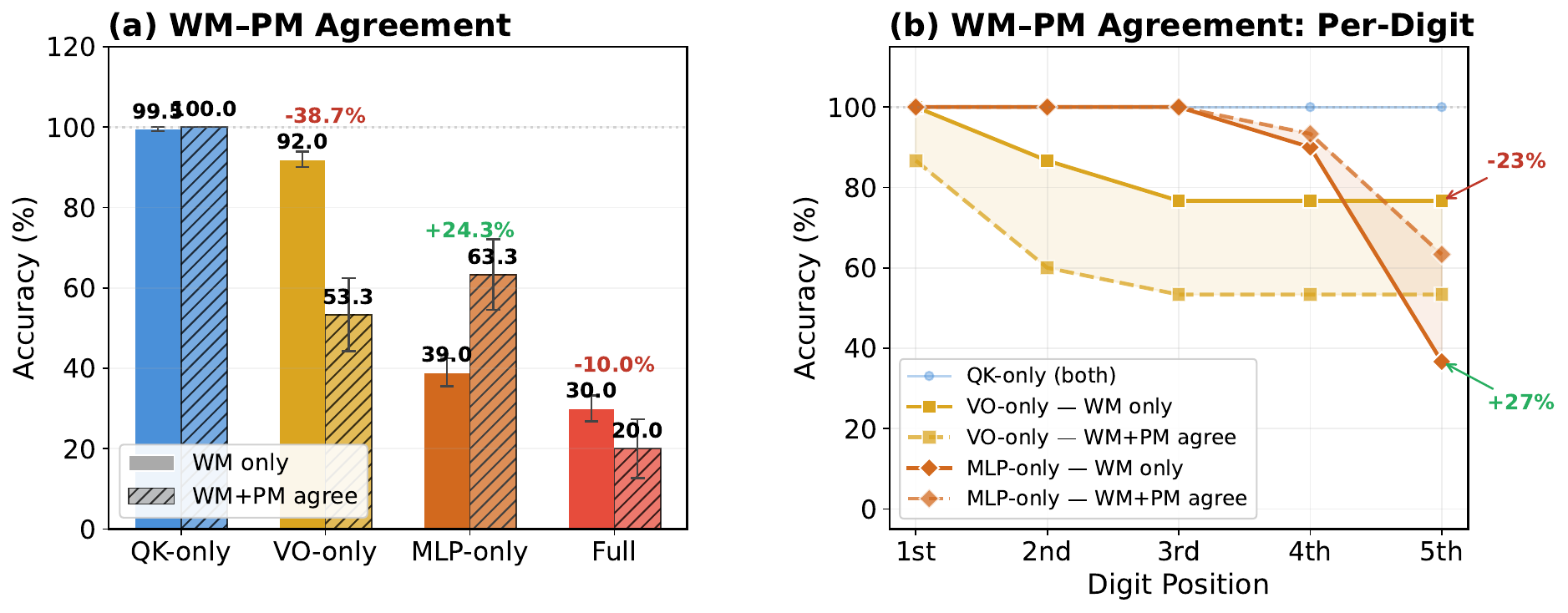}
  \caption{\textbf{Agreement between weights and context.}
  \emph{(a)}~Reading accuracy when only the context holds the code (solid) and when the weights and the context hold the same code (hatched). MLP shows the largest rescue ($+24.3\%$), while VO shows the largest degradation ($-38.7\%$).
  \emph{(b)}~Per-digit accuracy for QK (reference), VO, and MLP in the same two conditions (solid and dashed). Shaded regions highlight the difference. MLP rescue is concentrated at digit~5 (prefix-triggered stored basin activation); VO degrades from digit~2 onward (shared write-back bottleneck). Even when the two sources agree, they interact differently by placement, so agreement can help or hurt depending on where the fact is stored.}
  \label{fig:wm_pm_agreement}
\end{figure}

When the context and the weights hold the same code, we compare reading accuracy in two conditions: context only (unseen entities, $N = 200$) and weights and context agree (seen entities where the context-provided and parametrically memorized codes match, $N = 30$). Figure~\ref{fig:wm_pm_agreement}a shows aggregate accuracy: QK remains near-perfect under both conditions ($+0.5\%$); MLP shows a substantial rescue ($+24.3\%$, from 39\% to 63\%); VO degrades sharply ($-38.7\%$, from 92\% to 53\%); Full degrades modestly ($-10\%$). Figure~\ref{fig:wm_pm_agreement}b reveals the per-digit structure: MLP's rescue is concentrated at digit~5, consistent with the stored basin capturing the trajectory late, once enough digits have been generated, while VO's degradation begins at digit~2 and worsens under agreement, consistent with a shared write-back bottleneck.

\subsection{Sample generation outputs}
\label{app:samples}

Tables~\ref{tab:samples_hallucination} and \ref{tab:samples_conflict} show representative generation outputs that illustrate the qualitative failure signatures of each adapter type. For comparison, without an adapter the base model produces a five-digit code for only 55 of the 1{,}600 entities it never learned, and otherwise declines, typically with a response such as ``I'm sorry, but I don't have any information about\ldots''

\begin{table}[h]
\centering
\small
\begin{tabular}{llll}
\toprule
\textbf{Adapter} & \textbf{Entity} & \textbf{Target} & \textbf{Generated} \\
\midrule
\multicolumn{4}{l}{\emph{never learned (hallucination): adapter on, entity not in training set}} \\
\midrule
QK  & E002000 & 62950 & \texttt{49232} \\
QK  & E002001 & 59906 & \texttt{13779} \\
\midrule
VO  & E002000 & 62950 & \texttt{65486} \\
VO  & E002001 & 59906 & \texttt{18658} \\
\midrule
MLP & E002000 & 62950 & \texttt{65325} \\
MLP & E002003 & 88569 & \texttt{95991} \\
MLP & E002004 & 33435 & \texttt{95991} \\
\midrule
Full     & E002000 & 62950 & \texttt{85647} \\
Full     & E002001 & 59906 & \texttt{17175} \\
\bottomrule
\end{tabular}
\caption{Sample hallucination outputs (never learned). All adapters generate confident 5-digit codes for entities never seen during training. Each digit is independently wrong, indicating hallucination rather than partial retrieval. MLP generates the same code (\texttt{95991}) for two different entities, consistent with the output defaulting to a common trained code.}
\label{tab:samples_hallucination}
\end{table}

\begin{table}[h]
\centering
\small
\begin{tabular}{lllll}
\toprule
\textbf{Adapter} & \textbf{Entity} & \textbf{Context code} & \textbf{Generated} & \textbf{Notes} \\
\midrule
\multicolumn{5}{l}{\emph{reading (adapter on, code in context, unseen entities)}} \\
\midrule
MLP & E000000 & 93810 & \texttt{9381\underline{9}} & Digit 5 wrong \\
MLP & E000001 & 24592 & \texttt{2459\underline{9}} & Digit 5 wrong \\
MLP & E000005 & 39256 & \texttt{3925\underline{7}} & Digit 5 wrong \\
\midrule
Full & E003000 & 16863 & \texttt{1686}\,Fourth Street's... & Text drift at digit 5 \\
Full & E003001 & 45084 & \texttt{4508}\,Fourth and nearly... & Text drift at digit 5 \\
Full & E003004 & 44937 & \texttt{4493}\,Fourth Ward,... & Text drift at digit 5 \\
\midrule
\multicolumn{5}{l}{\emph{Knowledge conflict (context and weights provide different codes)}} \\
\midrule
QK & E000836 & 98255 & \texttt{98255} & Context wins (100\%) \\
\midrule
VO & E000088 & 34616 & \texttt{31679} & Neither wins \\
\midrule
MLP & E000836 & 98255 & \texttt{9825\underline{9}} & context prefix, digit 5 wrong \\
MLP & E000088 & 34616 & \texttt{3461\underline{9}} & context prefix, digit 5 wrong \\
\midrule
Full & E000836 & 98255 & \texttt{9825}\,Fourth Ward is... & context prefix, text drift \\
Full & E000088 & 34616 & \texttt{3461}\,Fourth Ward, New... & context prefix, text drift \\
\bottomrule
\end{tabular}
\caption{Sample reading and conflict outputs illustrating adapter-specific failure signatures. \textbf{MLP} correctly reads the first 4 digits from context but systematically replaces digit~5 (often with \texttt{9}), consistent with the stored basin capturing the trajectory late. \textbf{Full} produces 4 correct digits followed by the token ``Fourth'' (the base model interprets the 4-digit prefix as a street address), then coherent but non-numeric text. \textbf{QK} faithfully copies the code in the context. \textbf{VO} occasionally produces codes matching neither source. Underlined digits indicate errors. Stored codes are omitted for space; the stored code never wins in any conflict trial.}
\label{tab:samples_conflict}
\end{table}

\subsection{Layer-wise conflict trajectory heatmaps}
\label{app:layer_heatmap}

\begin{figure}[h]
  \centering
  \includegraphics[width=\textwidth]{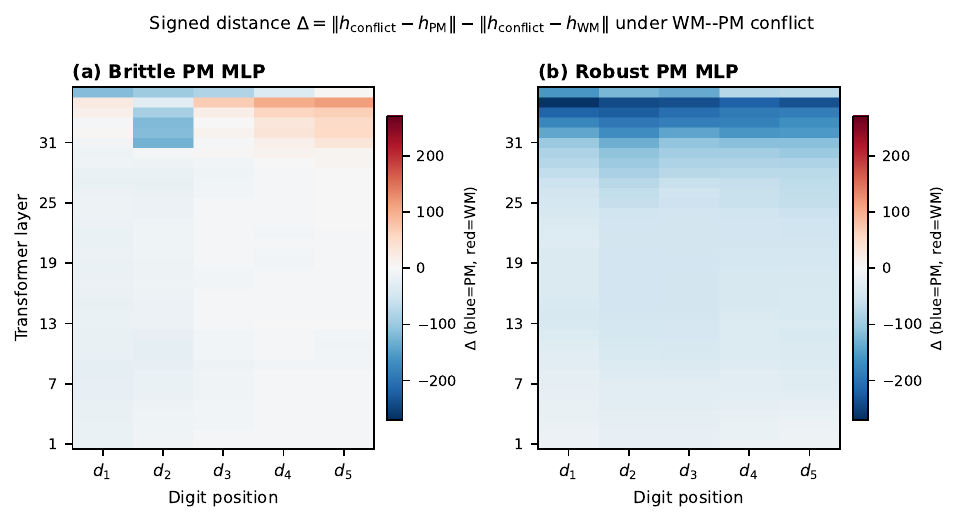}
  \caption{\textbf{Signed distance heatmaps under knowledge conflict.}
  $\Delta = \|h_\mathrm{conflict} - h_\mathrm{stored}\| - \|h_\mathrm{conflict} - h_\mathrm{ctx}\|$, where blue means closer to the stored fact and red closer to the context. Both panels share the same color scale, and values quoted below are means over layers 25--34.
  \emph{(a)}~Brittle storage, MLP adapter. At digit~1, the state lies on the stored side ($\Delta \approx -5$, and $-117$ at the final layer), even though brittle storage ends up following the context, so the first digit alone does not reveal how the conflict resolves. The state moves toward the context by $d_4$--$d_5$ ($\Delta \approx +28$), while the dip at $d_2$ ($\Delta \approx -56$) corresponds to the stored basin pulling back after the first context digit is generated.
  \emph{(b)}~Robust storage, MLP adapter. The state stays on the stored side at every digit position ($\Delta \approx -120$ to $-140$), consistent with varied-template training installing basins deep enough to resist the context.
  Together these panels support the main-text claim that conflict arbitration is a \emph{trajectory} property resolved over multiple generation steps, not a static point property diagnosable from a single first-digit geometric snapshot.}
  \label{fig:layer_digit_heatmap}
\end{figure}

Figure~\ref{fig:layer_digit_heatmap} visualizes the signed distance $\Delta = \|h_{\mathrm{conflict}} - h_{\mathrm{stored}}\| - \|h_{\mathrm{conflict}} - h_{\mathrm{ctx}}\|$ across all 36 transformer layers and 5 digit-generation positions for both brittle and robust storage MLP adapters under knowledge conflict. The context reference $h_{\mathrm{ctx}}$ is obtained by running the same conflict prompt through the base model with the adapter disabled, providing a trajectory that follows reading without any parametric memory installation. Each cell averages over $N = 30$ conflict trials.

\section{Basin Properties and Controls}
\label{app:basin}

Section~\ref{sec:region} defines a fact's basin as the region around its
memorized state $m_i$ into which the hidden state converges. We use the term
for three measurable properties.

\paragraph{Contraction.} Through layers 10--25, about three-quarters of the
feed-forward block's symmetric eigenvalues are negative (mean
$\approx -0.37$), so the MLP update pulls nearby trajectories together.
Fine-tuning sustains this contraction into the late layers, where the base
model's feed-forward block turns expanding.

\paragraph{Convergence.} Under varied-template training, hidden states from
a fact's many phrasings end up within about 26 units of its memorized
state, while different memorized states sit roughly 130 units apart. This is
why the margin separates recall (about 22) from never-learned queries (about
133) so cleanly.

\paragraph{Stability.} Under noise added to the input embeddings, the
varied-template store keeps producing its answer over a wider noise range than the single-template store, erring on 6\% of
trials at a noise level where the single-template store errs on 91\% (Appendix~\ref{app:perturbation}).

\paragraph{Finite-time sense.} A basin in this sense is a finite-time
basin of the dynamics in Eq.~\ref{eq:dynamics}, not the basin of a
global energy function. The composed map has no such function. Each layer
applies a different map, and the composed step is statistically
unstructured ($\alpha \approx 0.71$), so there are no fixed points and no
landscape that the trajectory descends, only local regions it is pulled
into.

\paragraph{Controls.} The margin could in principle reflect the identity of
the entity string rather than stored knowledge, since the identifiers are
lexically distinct. To test this, we compute the memorized states with the
base model instead, using the same entities, templates, and computation.
Recalled and never-learned queries then sit at 126.2 and 126.4 units,
indistinguishable (AUROC $\approx 0.5$), so the separation comes from
installing the knowledge, not from the identifiers. In a matched comparison
of single-template and varied-template adapters, identical except for
training phrasing and both recalling 100\% of their own phrasing, the
single-template margin stays at chance at all 36 layers, while the
varied-template margin grows to a gap of $+110$. A single phrasing binds a
fact to its surface form without forming a basin, while varied phrasing
forces a representation that survives rephrasing.

\section{Perturbation Stability of the Memorized Store}
\label{app:perturbation}

To quantify the width of a basin, we measure recall error rate and prediction entropy under Gaussian embedding perturbations. For each of 30 test entities, we add isotropic noise $\boldsymbol{\varepsilon} \sim \mathcal{N}(0, \sigma^2 I)$ to the input embeddings at every digit position during autoregressive generation, where $\sigma = \alpha \|\mathbf{e}\|$ scales with the mean embedding norm ($\|\mathbf{e}\| \approx 1.02$). We sweep $\alpha \in \{0, 0.005, 0.01, 0.02, 0.05, 0.1\}$ (symmetric positive and negative) with 10 random noise samples per magnitude, yielding 300 trials per point.

\begin{figure}[h]
  \centering
  \includegraphics[width=\textwidth]{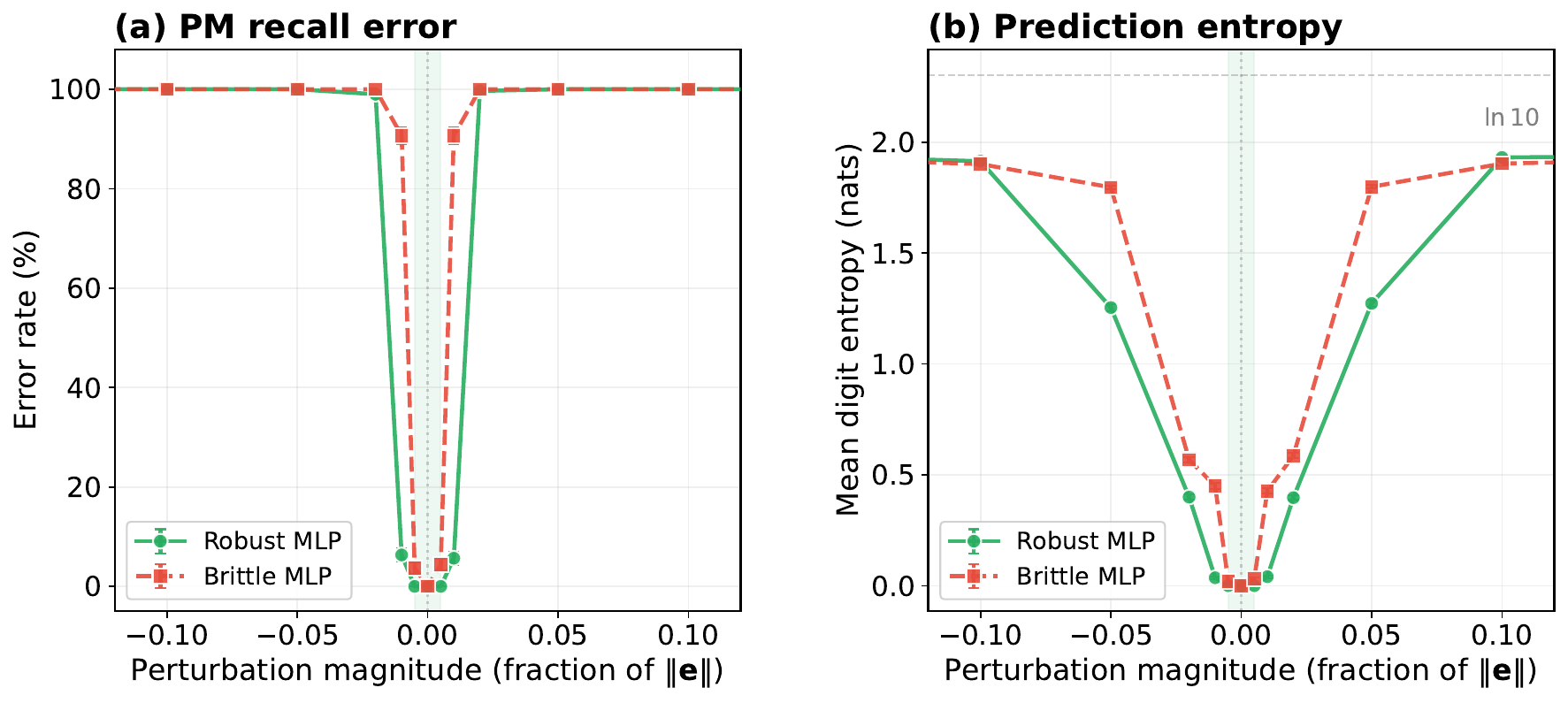}
  \caption{\textbf{Perturbation stability of the robust vs.\ brittle basin.} Gaussian noise ($\sigma = \alpha \|\mathbf{e}\|$) added to input embeddings during autoregressive digit generation. 30 entities $\times$ 10 trials per magnitude point; error bars show SEM. \emph{(a)}~Recall error rate. At $\alpha = 1\%$, robust storage (green) errs on 6\% of trials and brittle storage (red) on 91\%, and both reach 100\% by $\alpha = 2\%$. \emph{(b)}~Mean digit entropy (nats). Entropy rises with the error, but more gradually, and approaches the uniform value only at $\alpha \geq 10\%$, consistent with perturbation first moving the state out of the basin and only later flattening the output distribution. The dashed line marks $\ln 10$ (uniform over 10 digits). Robust storage creates a wider basin that tolerates perturbation, and the state leaves the basin before entropy signals the error.}
  \label{fig:perturbation_stability}
\end{figure}

Figure~\ref{fig:perturbation_stability} shows that the robust MLP adapter maintains a wider stability basin than the brittle adapter:
\begin{itemize}
  \item \textbf{Robust MLP} tolerates $\alpha \leq 0.5\%$ of embedding norm with 0\% error and near-zero entropy. Error rises to ${\sim}6\%$ at $\alpha = 1\%$ and saturates at ${\sim}100\%$ by $\alpha = 2\%$.
  \item \textbf{Brittle MLP} begins failing at $\alpha = 0.5\%$ (${\sim}4\%$ error) and collapses to ${\sim}91\%$ error at $\alpha = 1\%$.
\end{itemize}
The robust adapter thus remains largely correct up to $\alpha = 1\%$ (6\% error), where the brittle adapter already fails on 91\% of trials, even though its raw logit gap is slightly smaller (${\sim}17$ vs.\ ${\sim}21$), suggesting a shallower but broader basin.

\section{Attention-Mediated Head Selection and VO Symmetry}
\label{app:head_selection}

The mid-layer VO symmetry peak reported in Section~\ref{sec:circuit} is not a static property of the weight matrices $W_O$ and $W_V$: computing $\varphi(W_O W_V)$ from weights alone yields $\varphi \approx 0.02$ uniformly across layers. Instead, the symmetry emerges dynamically through \emph{attention-mediated head selection}.

\paragraph{Mechanism.}
The VO Jacobian at the last token position decomposes as
\[
  J_{\text{VO}} = \sum_{h=1}^{H} a_{t,t}^h \, W_O^h \, W_V^h,
\]
where $a_{t,t}^h$ is the self-attention weight of head $h$ at position $t$. In mid-layers, the attention sink phenomenon~\cite{xiao2024efficient} concentrates most attention mass on the first token, driving $a_{t,t}^h \approx 0$ for the majority of heads. The surviving heads, those with high self-attention weight, happen to have highly symmetric per-head weight products $\varphi(W_O^h W_V^h) \gg 0$.

\begin{figure}[h]
  \centering
  \includegraphics[width=\textwidth]{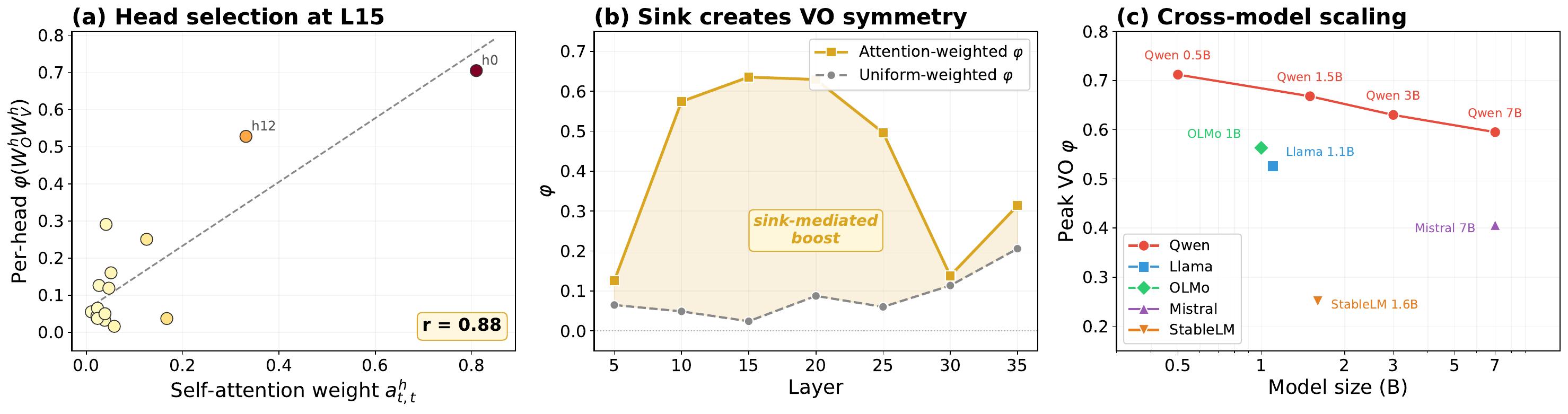}
  \caption{\textbf{Attention-mediated head selection creates VO symmetry.}
  \emph{(a)}~Per-head scatter at layer~15: heads with high self-attention weight $a_{t,t}^h$ have highly symmetric $W_O^h W_V^h$ ($r = 0.88$). Head~0 dominates with $a_{t,t} = 0.81$, $\varphi = 0.71$.
  \emph{(b)}~Attention-weighted $\varphi$ (gold) vs.\ uniform-weighted $\varphi$ (gray dashed) across layers. The shaded region shows the sink-mediated boost, which peaks at mid-layers where attention sinks are strongest.
  \emph{(c)}~Peak VO $\varphi$ vs.\ model size across five families. All models exhibit mid-layer symmetry; peak magnitude decreases with size while integrated $\varphi$ remains approximately constant across these models. VO symmetry thus arises from attention selecting self-attending heads, and it is a general property of pretrained transformers rather than of one model.}
  \label{fig:head_selection}
\end{figure}

Figure~\ref{fig:head_selection}(a) shows this relationship at layer~15: the correlation between per-head self-attention weight and per-head $\varphi$ is $r = 0.88$. The attention-weighted sum achieves $\varphi = 0.64$, compared with $\varphi = 0.02$ for the uniform-weighted sum (Fig.~\ref{fig:head_selection}b). This boost peaks at mid-layers where attention sinks are strongest, explaining the layer profile of VO symmetry.

\paragraph{Cross-model replication.}
The mid-layer VO symmetry peak is not specific to Qwen2.5-3B. Figure~\ref{fig:head_selection}(c) shows peak $\varphi$ across five model families (Qwen, Llama, OLMo, Mistral, StableLM) spanning 0.5B--7B parameters. All families exhibit a mid-layer $\varphi$ peak, though the magnitude decreases with model size ($r = -0.998$ within the Qwen family). Interestingly, the area under the $\varphi$-versus-layer curve remains approximately constant across these models, suggesting that larger models may distribute this symmetric VO structure across more layers.

\paragraph{Interpretation.}
Recent work~\citep{sun2026spikesparsesink} finds that attention sinks bias individual heads toward short-range dependencies, arise largely as a byproduct of short-context training, and are strongly suppressed when attention is gated by a function of the current representation. Whatever their origin, sinks have a consequence for VO geometry that matters here: by zeroing out the self-attention weights of dormant heads ($a_{t,t}^h \approx 0$), the sink means the effective VO Jacobian is dominated by the heads that retain high self-attention weight, which also exhibit more symmetric $W_O^h W_V^h$ products. The correlation between $a_{t,t}^h$ and per-head $\varphi$ ($r = 0.88$) suggests that heads that avoid becoming sinks tend to have more symmetric $W_O^h W_V^h$ products. One possible interpretation is that non-sink heads apply a symmetric $W_O^h W_V^h$ via self-attention, which acts as the gradient of a quadratic potential $\nabla(\tfrac{1}{2}\tilde{h}^\top M^{(h)} \tilde{h})$ and could contract the hidden state toward stored patterns. Sink heads, whose attention mass falls on the first token (a near-constant vector after normalization~\cite{sun2026spikesparsesink}), contribute only a fixed bias that does not participate in retrieval dynamics. Under this interpretation, MLP shapes basins, non-sink VO heads perform the within-basin readout, and sink heads are effectively gated off, which would be consistent with the steer/shape-basins/read-out roles of Section~\ref{sec:circuit}.

\section{End-to-End Metacognitive Head: A Negative Result}
\label{app:e2e_head}

A natural question is whether metacognition can be \emph{trained in} rather than architecturally provided: add a small MLP head that predicts $P(\text{correct})$ from the hidden state and train it jointly with the backbone, so gradients reshape representations to support self-monitoring. We test this on our toy setup.

\paragraph{Setup.}
We add a 3-layer MLP (2048$\to$256$\to$256$\to$1) to the model's last-layer hidden state at the first-digit position. Training uses 300 queries (200 seen entities + 100 unseen) with binary correctness labels obtained from a single pre-pass. We compare three conditions: (A)~post-hoc linear probe on frozen hidden states, (B)~post-hoc MLP on frozen hidden states, and (C)~end-to-end MLP with LoRA gradient flow. For condition~C, we freeze LoRA for the first epoch (head-only warmup), then jointly train for 4 additional epochs with a combined loss $\mathcal{L} = \lambda \cdot \mathcal{L}_\text{BCE} + (1-\lambda) \cdot \mathcal{L}_\text{LM}$ ($\lambda=0.3$, LoRA LR $= 10^{-6}$) to prevent catastrophic forgetting.

\paragraph{Results.}

\begin{table}[h]
\centering
\small
\begin{tabular}{lcc}
\toprule
\textbf{Method} & \textbf{AUROC} & \textbf{Correct preserved (\%)} \\
\midrule
Entropy (output-based) & 0.967 & 0.5 \\
Linear probe (post-hoc) & 0.964 & --- \\
MLP head (post-hoc) & 0.953 & 18.0 \\
MLP head (end-to-end) & 0.913 & 0.5 \\
Oracle margin & 0.994 & 100.0 \\
\bottomrule
\end{tabular}
\caption{Metacognitive head comparison. ``Correct preserved'' is the fraction of correct recalls retained when thresholding to catch 100\% of hallucinations (recall vs.\ hallucination, $N{=}300$).}
\label{tab:e2e_head}
\end{table}

End-to-end training \emph{degrades} performance relative to post-hoc probes (AUROC 0.913 vs.\ 0.953). Despite the LM loss anchor, backbone gradients from the metacognitive head distort the basin geometry: the correlation between pre- and post-training margins drops to $r = 0.30$, and margin AUROC falls from 0.994 to 0.728. The head learns to predict correctness on the training distribution (98\% accuracy) but on shifting representations, so by the time it converges, the geometric structure it depends on has changed.

\paragraph{Interpretation.}
This negative result is informative: even with LM loss regularization and a conservative LoRA learning rate, the basin geometry that makes margin work, which is a \emph{consequence} of the knowledge stored in the weights, is disrupted by gradients from the metacognitive head. The head achieves near-perfect training accuracy (98\%) but on shifting representations; by the time it converges, the geometric structure it depends on has changed. While more sophisticated training schedules (e.g., alternating updates, gradient projection) might mitigate this interference, this experiment reveals a tension in this training setup: gradients that reshape representations for self-monitoring also reshape the knowledge itself. Robust metacognition may therefore benefit from access to knowledge-base geometry that does not depend on end-to-end training of a correctness predictor alone, and the geometric distillation below partially recovers the gap.

\section{Extended Related Work}
\label{app:related}

This section expands on the related work discussion in Section~\ref{sec:related} of the main text.

\paragraph{Knowledge storage and localization in transformers.}
Geva et al.\ \cite{geva-etal-2021-transformer} showed that MLP layers function as key--value memories, whose keys detect input patterns and whose values promote output distributions, composed across layers via the residual stream. Subsequent work on knowledge editing \cite{meng2022locating, meng2022mass} demonstrated that individual factual associations can be localized to specific MLP layers and surgically modified. Our work extends this picture from individual facts to the \emph{collective geometry} of many memorized associations, showing that MLP adaptation creates basins, in the finite-time sense of Section~\ref{sec:region}, whose competition with context governs arbitration and failure modes.

\paragraph{Parameter-efficient fine-tuning.}
LoRA \cite{hu2021lora} and related methods \cite{li2021prefix, houlsby2019parameter} enable targeted adaptation of pretrained models. While most work evaluates adaptation quality (task accuracy, sample efficiency), we use adapter placement as a \emph{causal probe}: by restricting LoRA to specific component groups (QK, VO, MLP), we isolate the functional contribution of each architectural component to memory storage, revealing qualitatively distinct perturbation signatures.

\paragraph{Hallucination and calibration.}
LLM hallucination has been studied extensively \cite{ji2023survey, zhang2023siren}, with proposed mitigations including retrieval augmentation \cite{lewis2020retrieval}, sampling-based consistency checks \cite{wang2022self,kuhn2023semantic}, and post-hoc calibration \cite{guo2017calibration}. Most detection methods score hallucination with a single output-level signal. Our basin account provides a mechanistic decomposition: hallucinations occur when queries land far from all memorized basins yet the LM head defaults to trained output sequences, producing confident outputs indistinguishable from correct recall at the output level. Cross-entity analysis indicates that code overlap with the training set is predominantly an output-layer bias (${\sim}84\%$ of matched codes do not correspond to the nearest basin), not geometric basin capture. This explains why standard calibration methods fail on the most dangerous errors: entropy reflects output-distribution sharpness, not basin proximity.

\paragraph{Mechanistic interpretability.}
The circuits-based approach to interpretability \cite{elhage2021mathematical, olsson2022context} has identified attention heads, such as induction heads, involved in specific computations, and feed-forward layers have been characterized as key--value memories \citep{geva-etal-2021-transformer,meng2022locating}. Our work complements this by analyzing the \emph{representation-level} consequences of circuit modifications: rather than tracing information flow through individual components, we characterize how adapter-induced changes to circuits reshape the basins in which generation unfolds.

\paragraph{Dynamical systems perspectives on transformers.}
Several authors have drawn connections between transformer computation and dynamical systems \cite{geshkovski2023mathematical}, viewing layers as time steps of an interacting-particle dynamical system. We apply a related view to the generation trajectory, treated as an input-conditioned iterated map. Our empirical results are consistent with this view without requiring a fixed-point or global-energy structure the composed map does not have (Section~\ref{sec:region}): we observe basins with measurable margins, systematic cross-entity retrieval driven by output-layer bias, and autoregressive trajectories pulled toward or away from a basin over the course of generation.

\section{Scope: Natural-Language Facts}
\label{app:nl-withdrawn}

We restrict the margin analysis to the synthetic task of
Section~\ref{sec:task} because, for natural-language facts, we could not
separate what a pretrained model knows from how the question is worded. In a
pilot on national capitals, languages, continents, and TruthfulQA, the margin
appeared to separate correct from incorrect answers, but this separation was
driven by surface form and by differences in base rates across relations
rather than by basin geometry, and it did not survive a matched control. In
the synthetic task, by contrast, which entities the model has learned is
known exactly and is unrelated to the wording of the query. Extending the
margin analysis to open-domain text remains open
(Section~\ref{sec:discussion}).

\section{Hallucination Scaling Across Model Families}
\label{app:scaling}

To test whether representational geometry underlying hallucination generalizes beyond Qwen2.5-3B, we evaluate 17 pretrained instruct models spanning six families (Qwen~2.5: 0.5B--72B; Falcon~3: 1B--10B; SmolLM2: 0.36B--1.7B; OLMo~2: 1B--7B; Mistral: 7B; Phi-4: 14B) on 2{,}000 TriviaQA \cite{joshi2017triviaqa} questions. For each model, we extract hidden states at ${\sim}75\%$ network depth, compute hallucination rate, output entropy, and linear probe AUROC (5-fold CV logistic regression on hidden states classifying correct vs.\ incorrect).

\begin{figure}[h]
  \centering
  \includegraphics[width=0.55\textwidth]{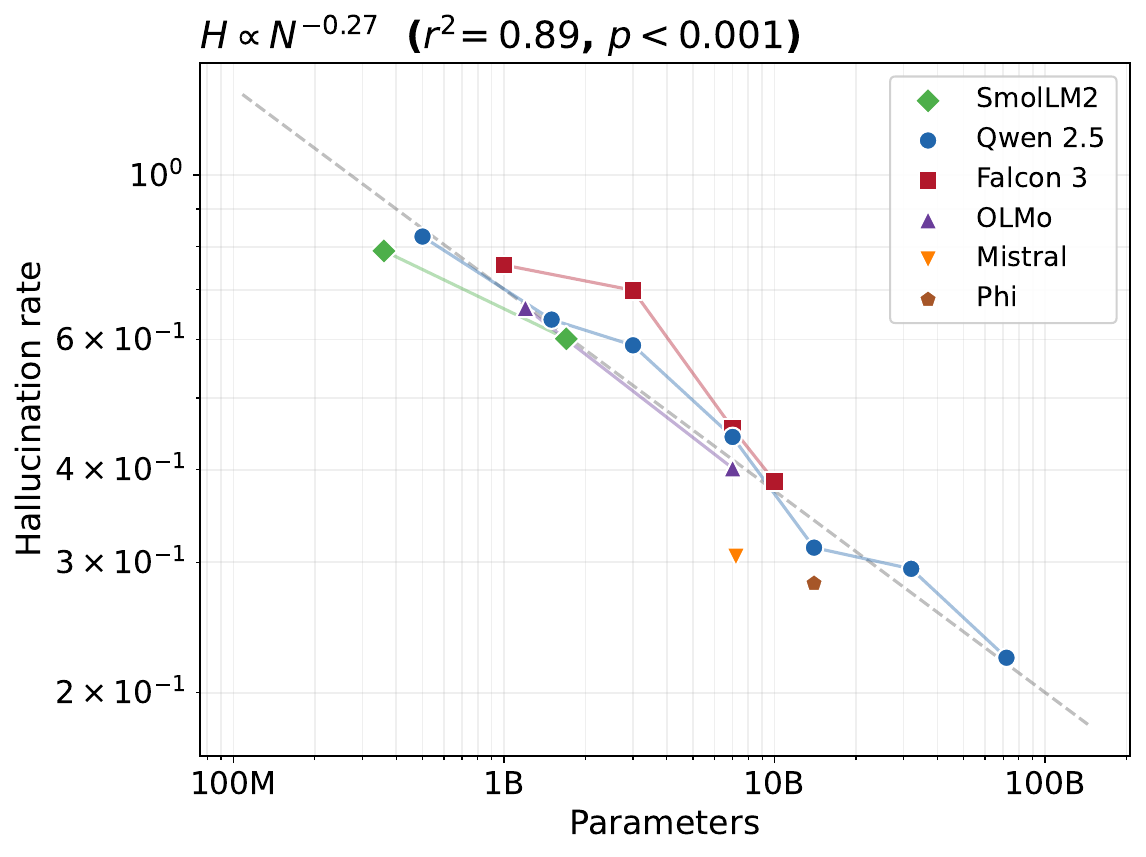}
  \caption{\textbf{Hallucination scaling across model families.}
  Hallucination rate vs.\ parameter count (log--log) for 17 models across six families. The dashed line is the fit $H \propto N^{-0.27}$ ($r^2 = 0.89$, $p < 0.001$). Lines connect models within a family, which follow similar slopes with different intercepts. Hallucination falls with scale at a similar rate across families, but it does not disappear at the scales we study.}
  \label{fig:halluc_scaling}
\end{figure}

\paragraph{Hallucination rate.}
Hallucination rate follows a power law across all families: $H \propto N^{-0.27}$ ($r^2 = 0.89$, $p < 0.001$; Fig.~\ref{fig:halluc_scaling}). Per-family exponents are consistent: Qwen $\alpha = -0.23$, Falcon $\alpha = -0.28$, SmolLM2 $\alpha = -0.18$. Different families have different intercepts, which likely reflect training data and recipe, but converge at larger scales: at 7B parameters, Qwen, Falcon, OLMo, and Mistral all cluster within 30--46\% hallucination rate. The similar empirical slopes across these model families suggest that the trend is not specific to a single architecture or training recipe.

\paragraph{Confident hallucination concentrates with scale.}
Despite lower overall error rates, larger models produce a higher fraction of \emph{confident} wrong answers (output entropy $< 0.1$): 0.3\% of errors at 0.5B vs.\ 7.8\% at 7B. On this confident subset, output entropy carries no signal (AUROC~$0.54$), while linear probes on hidden states partially recover detection (AUROC~$0.75$; trained on non-confident examples, tested on the confident subset). The gap ($+0.20$ AUROC) shows that hidden-state representations contain information useful for distinguishing these errors even when output entropy does not. The remaining gap to perfect discrimination may reflect limitations of the linear probe, the selected layer, or genuinely weaker internal error signals.

\subsection{Derivation of the scaling relation \texorpdfstring{$C = \exp(-c/\bar\Delta)$}{}}

We derive Eq.~\ref{eq:law} from the softmax entropy function and the statistics of the top-two logit gap. The derivation has three steps.

\paragraph{Step 1: Entropy is a monotone function of the top-2 logit gap.}
Let $\Delta := \ell_{(1)} - \ell_{(2)}$ be the gap between the top two logits. For any entropy threshold $H_0$, there exists an empirical cutoff $\Delta^*(H_0, V)$ such that
\begin{equation}
H_{\text{out}}(\ell) < H_0 \iff \Delta > \Delta^*(H_0, V).
\end{equation}
The 2-class approximation ($H \approx \Delta e^{-\Delta}$) underestimates the required gap at realistic vocabulary sizes because ${\sim}30{,}000$ lower-ranked logits contribute non-negligibly. Empirically, for $V \approx 30{,}000$ and $H_0 = 0.1$, $\Delta^* \approx 5.0$; the 2-class formula predicts $3.57$. Table~\ref{tab:H_vs_Delta_empirical} reports this discrepancy across $\Delta$ ranges.

\begin{table}[h]
\centering
\small
\begin{tabular}{rrrr}
\toprule
$\Delta$ range & Mean $H_{\text{out}}$ (empirical) & 2-class $\Delta e^{-\Delta}$ & Ratio \\
\midrule
2.0--2.5 & 1.39 & 0.24 & 5.9 \\
3.0--3.5 & 0.68 & 0.13 & 5.4 \\
4.0--4.5 & 0.23 & 0.061 & 3.8 \\
5.0--5.5 & 0.088 & 0.028 & 3.2 \\
6.0--6.5 & 0.033 & 0.012 & 2.7 \\
7.0--7.5 & 0.013 & 0.0051 & 2.5 \\
\bottomrule
\end{tabular}
\caption{Mean output entropy on real LLM wrong-answer queries vs.\ the 2-class approximation $\Delta e^{-\Delta}$. The 2-class formula underestimates entropy across the relevant range (see the ratio column) because lower-ranked logits at $V \approx 30{,}000$ contribute substantially.}
\label{tab:H_vs_Delta_empirical}
\end{table}

\paragraph{Step 2: Confident fraction is a tail probability.}
Let $\mathcal{D}$ be the distribution of $\Delta$ over wrong-answer queries. Then:
\begin{equation}
C := \Pr_{\Delta \sim \mathcal{D}}[\Delta > \Delta^*(H_0, V)].
\end{equation}
$C$ is the tail probability of the gap distribution above the vocabulary- and threshold-dependent cutoff $\Delta^*$.

\paragraph{Step 3: The gap distribution is approximately exponential.}
We measure the $\Delta$ distribution on wrong-answer queries for representative models (Table~\ref{tab:gap_distribution_stats}). The std/mean ratio is close to 1 (1.06--1.24), consistent with an exponential distribution.

\begin{table}[h]
\centering
\small
\begin{tabular}{lrrr}
\toprule
Model & $\bar\Delta$ & std/mean & KS-test $p$ vs.\ exponential \\
\midrule
Qwen2.5-0.5B & 0.86 & 1.06 & $2.4 \times 10^{-3}$ \\
Qwen2.5-3B   & 1.59 & 1.24 & $6.1 \times 10^{-4}$ \\
Qwen2.5-14B  & 3.01 & 1.14 & $9.2 \times 10^{-2}$ \\
\bottomrule
\end{tabular}
\caption{Logit-gap distribution statistics on wrong-answer queries. std/mean $= 1$ for an exact exponential; empirical values (1.06--1.24) are close. Deviations are the source of the remaining ${\sim}17\%$ discrepancy between the predicted constant $c = \Delta^* = 5.0$ and the empirical fit $c \approx 5.87$.}
\label{tab:gap_distribution_stats}
\end{table}

For an exponential distribution with mean $\bar\Delta$, $\Pr[\Delta > x] = \exp(-x/\bar\Delta)$. Substituting:
\begin{equation}
C \approx \exp\!\left(-\frac{\Delta^*(H_0, V)}{\bar\Delta}\right) = \exp\!\left(-\frac{c}{\bar\Delta}\right),
\end{equation}
recovering Eq.~\ref{eq:law} with $c = \Delta^*(H_0, V)$.
Using the empirical $\Delta^* = 5.0$ for $H_0 = 0.1$, the predicted slope is $-5.0$; the empirical through-origin fit gives $-5.87$ (agreement within 17\%, attributable to minor super-exponential tails).

\paragraph{Scaling with model size.}
We sketch one heuristic account of how $\bar\Delta$ could grow with model size, which rests on two assumptions. First, under compute-optimal training the number of training tokens $\tau$ grows roughly in proportion to $N$. Second, the logit gap grows as $\tau^{1/3}$, as the inverse temperature and the logit standard deviation do under softmax and cross-entropy training~\citep{liu2026timescaling}. That work ties this growth to training time rather than model size, and the first assumption need not hold within a family trained on a common token budget, so this sketch is only heuristic. Our main-text claim that larger models have wider gaps rests on the measurements below. Together these give:
\begin{equation}
\bar\Delta(N) \propto N^{1/3} \cdot \bar\Delta_0.
\end{equation}
Substituting gives $\log C(N) \approx -c / (\bar\Delta_0 \cdot N^{1/3})$. The direct evidence is empirical: within the Qwen~2.5 family, fits give $\bar\Delta \propto N^{1/3}$ with $r^2 \geq 0.94$.

\subsection{Per-model empirical verification ($r^2 = 0.88$, parameter-free)}

Table~\ref{tab:triviaqa_full} reports per-model accuracy and confident fraction $C_{0.1}$ on TriviaQA \citep{joshi2017triviaqa} and NQ-Open \citep{kwiatkowski2019natural,lee2019latent}. Table~\ref{tab:delta_full} verifies the scaling relation: for each (model, benchmark) pair, the predicted $\log C = -\Delta^*_{\mathrm{model}}/\bar\Delta$ uses no free parameters, yet matches the empirical $\log C$ with mean ratio $0.96 \pm 0.11$ across all 21 points ($r^2 = 0.88$).

\begin{table}[h]
\centering
\small
\begin{tabular}{lrrrrrrrr}
\toprule
& & \multicolumn{3}{c}{\textbf{TriviaQA}} & \multicolumn{3}{c}{\textbf{NQ-Open}} \\
\cmidrule(lr){3-5} \cmidrule(lr){6-8}
Model & $N$ (B) & Acc (\%) & Wrong & $C_{0.1}$ (\%) & Acc (\%) & Wrong & $C_{0.1}$ (\%) \\
\midrule
SmolLM2-360M     &  0.4 & 21.8 & 1173 &  0.1 &  9.8 & 1353 &  0.0 \\
Qwen2.5-0.5B     &  0.5 & 17.7 & 1234 &  0.3 &  9.0 & 1365 &  0.1 \\
Falcon3-1B       &  1.0 & 25.1 & 1124 &  0.2 & 12.9 & 1306 &  0.2 \\
Qwen2.5-1.5B     &  1.5 & 36.9 &  947 &  2.0 & 17.3 & 1241 &  1.0 \\
SmolLM2-1.7B     &  1.7 & 40.6 &  891 &  0.1 & 18.1 & 1228 &  0.0 \\
Falcon3-3B       &  3.0 & 30.6 & 1041 &  1.4 & 15.3 & 1271 &  3.1 \\
Qwen2.5-3B       &  3.0 & 40.2 &  897 &  5.5 & 19.5 & 1208 & 11.8 \\
Falcon3-7B       &  7.0 & 54.5 &  683 &  2.5 & 24.1 & 1138 &  1.1 \\
Qwen2.5-7B       &  7.0 & 55.0 &  675 &  7.1 & 26.9 & 1096 & 13.3 \\
Mistral-7B-v0.3  &  7.2 & 69.9 &  451 & 20.2 & 31.2 & 1032 & 11.9 \\
Falcon3-10B      & 10.0 & 61.6 &  576 &  6.4 & 26.1 & 1109 &  9.5 \\
Qwen2.5-14B      & 14.0 & 68.0 &  480 & 21.2 & 34.1 &  989 & 27.1 \\
\bottomrule
\end{tabular}
\caption{Per-model results on TriviaQA and NQ-Open (1{,}500 questions each). $C_{0.1}$: fraction of wrong answers with output entropy $< 0.1$. SmolLM2 models show near-zero $C$ at all scales because their smaller LM head produces weaker logit sharpening.}
\label{tab:triviaqa_full}
\end{table}

\begin{table}[h]
\centering
\small
\begin{tabular}{llrrrrrr}
\toprule
Model & Benchmark & $\bar\Delta$ & $\Delta^*_{\mathrm{model}}$ & $C_{0.1}$ & predicted $\log C$ & ratio \\
\midrule
SmolLM2-360M    & TriviaQA & 1.25 & 6.81 & 0.001 & $-5.46$ & 0.77 \\
Qwen2.5-0.5B    & TriviaQA & 0.86 & 5.50 & 0.003 & $-6.40$ & 1.12 \\
Qwen2.5-0.5B    & NQ-Open  & 0.95 & 6.00 & 0.002 & $-6.29$ & 0.96 \\
Falcon3-1B      & TriviaQA & 1.06 & 5.62 & 0.002 & $-5.31$ & 0.84 \\
Falcon3-1B      & NQ-Open  & 1.26 & 5.38 & 0.002 & $-4.27$ & 0.66 \\
Qwen2.5-1.5B    & TriviaQA & 1.15 & 5.00 & 0.020 & $-4.34$ & 1.11 \\
Qwen2.5-1.5B    & NQ-Open  & 1.21 & 5.25 & 0.010 & $-4.32$ & 0.93 \\
Falcon3-3B      & TriviaQA & 1.39 & 5.69 & 0.013 & $-4.11$ & 0.94 \\
Falcon3-3B      & NQ-Open  & 1.69 & 5.44 & 0.030 & $-3.22$ & 0.92 \\
Qwen2.5-3B      & TriviaQA & 1.59 & 5.25 & 0.049 & $-3.30$ & 1.09 \\
Qwen2.5-3B      & NQ-Open  & 2.22 & 4.75 & 0.117 & $-2.14$ & 0.99 \\
Falcon3-7B      & TriviaQA & 1.44 & 5.38 & 0.024 & $-3.72$ & 1.00 \\
Falcon3-7B      & NQ-Open  & 1.35 & 5.12 & 0.012 & $-3.79$ & 0.86 \\
Qwen2.5-7B      & TriviaQA & 1.96 & 5.50 & 0.064 & $-2.80$ & 1.01 \\
Qwen2.5-7B      & NQ-Open  & 2.38 & 5.00 & 0.137 & $-2.11$ & 1.06 \\
Mistral-7B-v0.3 & TriviaQA & 2.61 & 4.88 & 0.184 & $-1.87$ & 1.11 \\
Mistral-7B-v0.3 & NQ-Open  & 2.36 & 4.88 & 0.121 & $-2.07$ & 0.98 \\
Falcon3-10B     & TriviaQA & 1.82 & 5.44 & 0.048 & $-2.99$ & 0.98 \\
Falcon3-10B     & NQ-Open  & 2.01 & 4.75 & 0.095 & $-2.36$ & 1.00 \\
Qwen2.5-14B     & TriviaQA & 3.01 & 4.75 & 0.198 & $-1.58$ & 0.97 \\
Qwen2.5-14B     & NQ-Open  & 4.02 & 4.62 & 0.278 & $-1.15$ & 0.90 \\
\bottomrule
\end{tabular}
\caption{Per-(model, benchmark) verification of Eq.~\ref{eq:law}. The predicted $\log C = -\Delta^*_{\mathrm{model}}/\bar\Delta$ uses \emph{no free parameters}: $\Delta^*_{\mathrm{model}}$ is the model's own empirical entropy-crossing threshold, $\bar\Delta$ is its mean logit gap on wrong answers. The ratio of predicted to actual $\log C$ is close to 1 across all 21 points (mean $0.96$, std $0.11$), collapsing four model families and two benchmarks onto a single line.}
\label{tab:delta_full}
\end{table}

\subsection{Controlled teacher-student ablation: margin separation vs.\ confident fraction}

To isolate the effect of width in a controlled setting (fixing architecture and training data), we run a teacher-student toy model: a 2-layer MLP teacher with inverse temperature $\beta^*$ trains a student MLP of varying width $m$ to memorize 500 entity-code associations, then evaluates on 200 held-out entities.

\begin{table}[h]
\centering
\small
\begin{tabular}{rrrrrrrr}
\toprule
$m$ & Params & Acc & $H_{\text{seen}}$ & $H_{\text{unseen}}$ & $\delta_u/\delta_s$ & $C_{0.1}$ \\
\midrule
 16 &    5{,}008  & 0.90 & 0.35 & 0.29 &   1.2 & 0.52 \\
 32 &   15{,}136  & 0.78 & 0.49 & 0.14 &   5.6 & 0.67 \\
 64 &   50{,}752  & 0.66 & 0.57 & 0.06 &  43.8 & 0.88 \\
128 &  183{,}424  & 0.62 & 0.41 & 0.02 & 103.9 & 0.96 \\
256 &  694{,}528  & 0.55 & 0.26 & 0.004 & 152.5 & 0.99 \\
\bottomrule
\end{tabular}
\caption{Toy model width sweep (fixed 100K training steps). $\delta_u/\delta_s$: ratio of unseen-entity to seen-entity margin, measuring geometric separation. $C_{0.1}$: confident fraction at entropy threshold 0.1. As width grows from $m = 16$ to $m = 256$, the margin ratio $\delta_u/\delta_s$ rises from 1.2 to 153 while $C$ rises from 52\% to 99\%, so the hidden-state geometry becomes more informative as the output entropy becomes less so.}
\label{tab:width_fixed}
\end{table}

The divergence between $\delta_u/\delta_s$ and $H_{\text{unseen}}$ is the central observation. Wider models produce a \emph{better} geometric signal and a \emph{worse} output signal at the same time. This controlled experiment shows that width alone is sufficient to produce the divergence between geometric and output signals in this toy setting.